%% file: main.tex
\documentclass[usenames,dvipsnames]{article}
\usepackage[dvipsnames]{xcolor}

\usepackage[english]{babel}
\usepackage[super]{nth}

\usepackage[letterpaper,top=2cm,bottom=2cm,left=3cm,right=3cm,marginparwidth=1.75cm]{geometry}

\usepackage{amsmath}
\usepackage{graphicx}
\usepackage[colorlinks=true, allcolors=blue]{hyperref}
\usepackage{array}
\usepackage{multirow}
\usepackage{booktabs}
\usepackage{pdflscape}
\title{What Can Secondary Predictions Tell Us?\\ An Exploration on Question-Answering with SQuAD-v2.0}

\author{Michael Kamfonas \\
  \texttt{mkamfonas@infokarta.com} \\
  \\
  Gabriel Alon \\
  \texttt{galon@umich.edu} 
}

\begin{document}
\maketitle
\input{00A_Abstract}

\input{10A_Introduction}

\input{30A_Methods}
\input{60A_Experiments}

\input{70A_Discussion}
\bibliographystyle{alpha}
\bibliography{references}
\newpage
\appendix
\input{35A_Models}

\input{37A_Metrics} 
\input{A20_Examples}

\begin{landscape}
\input{A21_ExamplePredictions4}
\end{landscape}

\input{A30_PolarizedExps}
\input{A32_PolarizedExps}

\end{document}

%% file: 00A_Abstract.tex
\begin{abstract}
Performance in natural language processing, specifically for the question-answering task, is typically measured by comparing a model's most confident (primary) prediction to  golden answers (i.e., the ground truth). We are making the case that it is also helpful to quantify how close a model came to predicting a correct answer for examples that failed, a goal that the F1 score only partially satisfies. We derive our metrics from the probability distribution from which the model ranks its predictions. We define an example's Golden Rank (GR) as the rank of its most confident prediction that matches the ground truth. Thus the GR quantifies how close a correct answer comes to being a model's best prediction.  We demonstrate how the GR can be used to classify questions and visualize their spectrum of difficulty, from relative successes to persistent extreme failures.

We derive a new aggregate statistic, the Golden Rank Interpolated Median (GRIM), that quantifies the proximity of correct predictions to the model's choices over a whole dataset or any slice of examples. To develop some intuition and explore the applicability of these scores, we use the \emph{Stanford Question Answering Dataset (SQuAD-2)} and a few popular transformer models from the Huggingface hub. We demonstrate that the GRIM is relatively independent of the F1 or the exact match (EM) scores. We then calculate and visualize these scores for various transformer architectures, probe their applicability in error analysis or difficulty assessment, and see how they relate to standard training diagnostics, i.e., the EM and F1 scores. We finally suggest  possible follow-up areas of research. 
\end{abstract}

%% file: 10A_Introduction.tex
\section{Introduction}

\input{12A_Overview}

\input{14A_Literature}

%% file: 12A_Overview.tex
\subsection{Motivation}
In the course of studying the NLP extractive question-answering task with various transformer-based models on the SQuAD v2.0 dataset \cite{rajpurkar2018know}, we became interested in understanding secondary prediction behavior. Specifically, we wanted to know how close correct answers came to becoming predictions and to use this information to  classify failures or assess example difficulty. Our notion of proximity differs from the way F1 measures approximate success. Our idea is to employ a rank-based metric, ordering predictions by descending confidence level (the most confident at rank 0) and quantifying proximity by the rank of the lowest-ranking correct prediction. Examples successfully predicted would rank 0; all others would rank higher. 

The primary validation method in  SQuAD \cite{rajpurkar2016squad} is the exact match test (EM), a rigid metric that compares predicted answers to a short list of annotated correct answers after normalizing the compared strings.
Normalization entails standardizing white space and eliminating punctuation and articles. The F1 score is a softer metric that rewards partial success by matching individual words rather than complete answers whenever the EM criterion fails.  

Although the F1 is appropriate as an extension of the EM, it is problematic when used for evaluating approximate success. Some of the issues are: 

\begin{enumerate}    
    \item The F1 is based on the top choice (primary prediction) and disregards  secondary predictions.
    \item It lacks proportionality, particularly with SQuAD-v2 when unanswerable questions are involved.
    \item It sometimes rewards answers that a human would find unacceptable 
\end{enumerate}

The high degree of overlap between the F1 and EM results in an almost linear relationship, as shown visually in figure \ref{M:EM_F1_CORR}. 
The F1 may only differ from the EM for responses that fail the exact match test and are longer than one word. 
There are also cases where F1's partial credit may reward incorrect answers or, particularly in SQuAD-2, give no credit to near-misses. We consider near-misses to be predictions matching golden answers that, although they don't make the primary choice, still rank close to it. 

For example, take this passage fragment which contains the  golden answer, highlighted in green: \emph{Victorian lines mainly use the \textcolor{OliveGreen}{1,600 mm (5 ft 3 in) broad gauge}.} Since ``1,600" is roughly equivalent in units to ``5 ft 3 in", there are multiple legitimate ways to answer the same question. An additional answer given in the dataset is: \emph{\textcolor{OliveGreen}{1,600 mm}}. Additional answers could be construed as correct, such as  \emph{\textcolor{red}{1,600 mm (5 ft 3 in)}}  or \emph{\textcolor{red}{5 ft 3 in}} or stretching it 
 \emph{\textcolor{red}{(5 ft 3 in) broad gauge}} which the F1  would, at least partially, reward even though no golden answer is explicitly provided. However, a prediction that returns \emph{broad gauge} would be incorrect, although the F1 will give it partial credit. An  answer such as \emph{mm (5 ft 3 in)} is hard to justify any partial credit for since the inclusion of ``mm" is nonsensical, but the F1 would reward it.
 
 As an  example of lack of proportionality, consider a case where the most confident prediction (rank-0) is ``No Answer" while  rank-1 contains the correct answer. The F1 score would give no credit, but the rank in predictive probability order would be 1 (second best). In our judgment, the model came very close to choosing the correct answer, i.e., one step away from the top prediction. Furthermore, in the presence of ``no-answers", two failed predictions have no way of being compared using the F1 test. 
 The rank-based method manifests ``proximity to the correct answer" in a proportional way that parallels the  mechanism used to choose the best answer.
 Another way of looking at this is as if the F1 score says ``there are no correct words in the answer," while the rank-based method tells us: ``the model's second choice was correct."

We propose a metric we call the \emph{Golden Rank (GR)}. We order an example's list of predictions by descending probability and find the predictions that match any of the correct answers. The lowest rank of those matched is the golden rank for the example. We also define an aggregate statistic we call \emph{Golden Rank Interpolated Median (GRIM.)}. The GRIM is calculated over secondary GRs only and is an estimator of  the proximity between golden answers and the top prediction for a sample of examples. Two experiments with the same EM score can have significantly different GRIM scores indicating that for the questions that failed, one model is consistently  assigning higher (alas, not the highest) confidence to the correct predictions than the other.

%% file: 14A_Literature.tex
\subsection{Related Work }

The SQuAD datasets  \cite{rajpurkar2016squad}, \cite{rajpurkar2018know}  support the extractive \emph{Question-Answering task} and provide machine learning models good quality large scale data to learn from with standardized methods of evaluation. 
The history of question-answering datasets traces back to TREC-8,  ``the first large-scale evaluation of domain-independent
question answering systems" \cite{voor1999TREK8}. The proposed Mean Reciprocal Rank (MRR) can assess multiple relevant results to a web search, which returns multiple predictions in response to a question. The question-answering task defined by SQuAD evaluates only the primary (best) prediction from a model with  the EM and F1 scores as the established metrics.

Secondary predictions, i.e.,  predictions that fail conventional model evaluation,  are  used primarily in error analysis and interpretability studies. These goals have led to hybrid human-machine approaches with implementations such as Pandora \cite{nushi2018towards}. The premise that there is no silver bullet in analyzing wrong predictions or interpreting the reasons for failure led to an elaborate suite of tools made available to aid the researcher. These tools help in analyzing errors, clustering them, categorizing them, and deriving various visual and diagnostic instruments to tackle families of failures in a more efficient machine-assisted manner.  

Radev et al. in \cite{evalQAsys2002} describe early rank-based metrics that can handle multiple predictions. They include, among others, First Hit Success (FHS), First Answer Reciprocal Word Rank (FARWR), Total Reciprocal Rank (TRR), and First Answer Reciprocal Rank (FARR).  The discounting effect of reciprocal rank, as in the Mean Reciprocal Rank (MRR) \cite{voor1999TREK8}, is to reward answers closer to the best prediction and penalize those that are further away in a concise metric that reflects the perceived utility in  the document search. 

When it comes to evaluating the levels of difficulty of examples in a dataset, a systematic framework, ILDAE, was proposed by \cite{varshney-etal-2022-ildae}, which  supports five discrete purposes:

\begin{enumerate}
    \item Efficient model evaluation by reducing the size of datasets in model comparisons
    \item Improving dataset quality by enhancing trivial examples and repairing erroneous ones
    \item Model analysis by difficulty class, aiding in the selection of models for particular situations
    \item Projection of out-of-domain (OOD) generalization potential by the use of difficulty-weighted accuracy
    \item Dataset difficulty analysis informing future dataset development
\end{enumerate}

The authors use a RoBERTa-large transformer
model for classifying each example by difficulty
level. The idea is to perform the specific processing tasks involved in difficulty evaluation once,
tagging instances accordingly. Then they compare other models, trained over subsets drawn over
prescribed difficulty distributions. These layered
subsets produce evaluation results showing a high-ranked correlation (Kendall) with those
trained on the complete datasets.
The authors claim substantial improvements in accuracy scores by modifying too trivial examples and
correcting potential errors that make particular examples too tricky. Similar insights
may inform the construction of altogether new
datasets.

According to \cite{https://doi.org/10.48550/arxiv.2205.09898}, example and dataset-level difficulty scoring can also facilitate curriculum learning strategies \cite{xu-etal-2020-curriculum} for multi-task learning (MTL) by enabling models to form their own curriculum. The methodology also reduces the high computational cost of automated dataset curriculum-forming methods and removes the uncertainty introduced when human judgment  determines the ordering. Finally, when applied to low-data regimes, the methodology is most effective for difficult examples, making it more appropriate for real-world applications.

Pan et al. in \cite{PanQASQuAD2} attempted to define a loss function based on how far away a false positive is from the ground truth. The notion of distance used was between the false positive and ground truth in the passage. It may be possible to apply the same idea, only expressing distance as rank, logit, or probability difference of the same end-points.

%% file: 30A_Methods.tex
\section{Methods}

\input{36A_Secodary_Metrics}
\label{sec:30A}



%% file: 36A_Secodary_Metrics.tex
\input{threeRandTopKpredictions_edited}
A primary prediction is the model's best choice of the span of words from the context\footnote{the passage associated with an example providing the context for answering the question} that answers the question. All possible spans, including the empty span,  comprise an ordered sequence that starts with the primary prediction followed by all secondary predictions from most likely to least likely. Because all golden answers are valid spans, each has at least one match in the sequence.  A golden answer may match multiple spans since articles and punctuation tokens get removed before the comparison. For example, the same golden answer may match  ``the Amazon" as well as ``Amazon." Answers that don't match any prediction span may signify  dataset errors. 

We refer to these matches as correct predictions.  We define the \emph{Golden Rank} (GR) of an example as the lowest rank of all correct predictions. If the Golden Rank is 0, i.e., the primary prediction, it scores as an exact match. So the ratio of the count of primary predictions to the total number of examples is the EM-score.

  Appendix \ref{L2P} describes in detail the preprocessing transformations from logits to arrive at these predictions. 

Let the top-K probability predictions for the  $i^\text{th}$ example be $\hat{y}^{A(i)}_{r}$, where $\hat{y}$ is a string and stands for the prediction at rank $r$. The superscript $A$ stands for \emph{answer}. For each of these top-K answers, the corresponding probabilities are $\hat{p}^{(i)}_r$. The golden rank $r_G^{(i)}$ is the rank of the most likely exact match to any of the $G^{(i)}$ golden answers for the example:

\begin{equation}
    \text{GR}:\,\,\,\, r_G^{(i)} = \underset{r\in [0,K]}{\arg\max}\,(\hat{p}^{(i)}_r|\hat{y}^{A(i)}_{r}=y^{A(i)}_j\, \forall{j \in G^{(i)}})
    \label{eq:GRIM}
\end{equation} 

In practice, extracting  all ranks of each example for each experiment would be grossly cost-prohibitive. A more reasonable compromise is to produce an output with each example's  best-$K$ list from which we can calculate the GR. We use $K=10$ for all our experiments here. So examples ranking higher than $K$ are assigned GR$\,=K$.  

Table \ref{tab:topKpreds} contains the top-10 predictions for three examples. It shows the rank, the predictions, the golden answers, the probability, and whether the prediction is correct. The first and second examples match at rank 0, i.e., the primary prediction. The primary prediction of the third example came out as no answer, which does not match the correct answer; however, predictions at ranks 1, 2, and 4 of the same example are correct. The Golden Rank for this last example is 1, i.e., it is a near miss, but it fails the EM criterion, and the F1  gives it no partial credit.

We also defined an aggregate score, the Golden Rank Interpolated Median (GRIM), over secondary predictions, i.e., examples with $GR > 0$. We considered various alternatives before settling on the GRIM: 

\begin{description}
    \item[The Golden Rank Mean (GRM)] is the mean of all golden ranks, which would be robust to different context lengths, but it requires that we know the actual golden ranks for all examples. Since we only collect the top-K predictions, only ranks $[0, K-1]$ reflect their true  GRs, while the remaining outliers are all assigned the same catch-all rank $K$, which distorts the mean.
    \item[The Discounted Golden Rank Mean (DGRM)] offers a way to avoid the mean outlier problem by discounting the GR contribution as ranks $r$ increase. A  factor such as $r^{-\gamma}$ with an appropriate value of $\gamma$ can make contributions higher than K insignificant. An implicit assumption is that golden answers that match lower ranks are of more interest. 
    \item[The Golden Rank Interpolated Median (GRIM)] is robust to fluctuations of K, as long as $K>\frac{N}{2}$ where N is the number of candidate spans. A pure median would only yield an integer since ranks are integers. The \href{https://en.wikipedia.org/wiki/Median#Interpolated\_median}{interpolated median} treats example ranks as midpoints of continuous intervals and assumes linear interpolation. If the regular median is $m$, and there are $k$ examples that fall above the median rank, $j$ examples in the median rank, and $i$ examples below the median rank, then the interpolated median\footnote{definition from Wikipedia} is given by

    \begin{equation}
         m_\text{interpolated} = m - \frac{1}{2} \left[\frac{k - i} j\right]
         \label{eq:intermedian}
    \end{equation} 

    \item[The First Answer Reciprocal Rank (FARR)] is described by Radev et al. in  \cite{evalQAsys2002}, and it is similar to the DGRM metric defined earlier with $\gamma=1$. For example, for rank 3, the FARR $= \frac{1}{3}$. 
    
    \item[The Mean Reciprocal Rank (MRR)] is the mean of the  inverse of the ranks of  potentially multiple correct values \cite{evalQAsys2002}.  It is well suited to evaluate the normalized performance of searches that return many answers. 
\end{description}
We chose to avoid discounting or normalizing ranks in our GR aggregation score implemented here, so we decided on the GRIM. 
We  use the vector of golden ranks produced by \autoref{eq:GRIM} to calculate the median $\tilde{r}_G$, and the counts $k$, $i$ and $j$ used in the interpolation equation \ref{eq:intermedian}, which computes the GRIM. 

%% file: threeRandTopKpredictions_edited.tex
\begin{table}
    \centering
    \title{\Large{Examples of Ranked Predictions}}
\begin{tabular}{| m{0.3cm} | m{0.6cm} | m{18em} | m{9em} | m{2cm} | m{1cm} |}
\hline
                      id &  Rank & Predicted Answer &   Golden Answers &   Probability &  Match \\
\hline
8b &     0 &                                                    &                    \multirow{10}{15em}{[]} &  1.000e+00 &     True \\
8b &     1 &                                            far off &                                &  4.217e-08 &    False \\
8b &     2 &                       in their own city or far off &                                &  1.025e-08 &    False \\
8b &     3 &  Galaxy Public School in Kathmandu. Most of the... &                                &  9.435e-09 &    False \\
8b &     4 &                                      Public School &                                &  2.582e-09 &    False \\
8b &     5 &  Public School" appended to them, e.g., the Gal... &                                &  9.224e-10 &    False \\
8b &     6 &  far off, like boarding schools. The medium of ... &                                &  9.001e-10 &    False \\
8b &     7 &                                           far off, &                                &  6.632e-10 &    False \\
8b &     8 &                                  un-aided' schools &                                &  4.343e-10 &    False \\
8b &     9 &    'aided' schools. The private 'un-aided' schools &                                &
3.816e-10 &    False \\
\hline
e1 &     0 &                                        vertebrates &  \multirow{10}{15em}{early vertebrates\\ vertebrates} &  9.999e-01 &     True \\
e1 &     1 &                                          rtebrates &   &  4.125e-05 &    False \\
e1 &     2 &                                             brates &   &  7.859e-06 &    False \\
e1 &     3 &                                                 ve &   &  2.296e-06 &    False \\
e1 &     4 &                                  early vertebrates &   &  6.034e-07 &     True \\
e1 &     5 &                                              verte &   &  2.536e-07 &    False \\
e1 &     6 &  adaptive immune system evolved in early verteb... &   &  1.369e-08 &    False \\
e1 &     7 &  The adaptive immune system evolved in early ve... &   &  4.158e-09 &    False \\
e1 &     8 &                                                rte &   &  1.046e-11 &    False \\
e1 &     9 &                                           early ve &   &  1.385e-12 &    False \\

\hline
c0 &     0 &                                                    &  \multirow{10}{15em}{about twice as much \\twice as much\\ twice\\ 50\% more} &  9.903e-01 &    False \\
c0 &     1 &                                      twice as much &   &  5.086e-03 &     True \\
c0 &     2 &                                              twice &   &  4.352e-03 &     True \\
c0 &     3 &                        twice as much (14.6 mg·L-1) &   &  1.616e-04 &    False \\
c0 &     4 &                                about twice as much &   &  3.260e-05 &     True \\
c0 &     5 &                                        about twice &   &  2.790e-05 &    False \\
c0      6 &     &                    twice as much (14.6 mg·L-1 &   &  6.369e-06 &    False \\
c0 &     7 &                             twice as much (14.6 mg &   &  5.840e-06 &    False \\
c0 &     8 &  twice as much (14.6 mg·L-1) dissolves at 0 °C ... &   &  2.609e-06 &    False \\
c0 &     9 &                  about twice as much (14.6 mg·L-1) &   &  1.035e-06 &    False \\
\hline

\end{tabular}
\caption{Top 10 predictions and golden answers for three examples. Rank 0 denotes the primary prediction, all other ranks are secondary predictions.  The Match column identifies whether the prediction exactly matches a golden answer}
\label{tab:topKpreds}
\end{table}

%% file: 60A_Experiments.tex
\section{Experiments}

We performed experiments to answer the following questions. 
\begin{enumerate}
    \item What is the Relationship among GR/GRIM, EM, and F1? How correlated are these metrics?
    \item What do the GR and the GRIM tell us about the models?
    \item What do the GR and the GRIM tell us about the examples, their difficulty level, and their  error patterns?
    \item How do the GRIM, EM, and F1 change during training?
    \item Is the GRIM, the EM, or the F1 a better criterion for selecting model ensembles?
\end{enumerate}
\begin{table}
    \centering
    \begin{tabular}{l|c|r|r|r|r}
    \hline
     Model/Experiment & Previously & +FT Epochs & Batch & Initial & Decayed \\
     Description & Fine-Tuned & (Sample Size) & Size &   LR & Opt Steps \\
\hline

       BERT-base-uncased-001 & - & 8 & 24 & 5e-5 & 43920\\
       BERT-base-uncased-007 & - & 8 & 24 & 3e-5 & 43920\\
       BERT-base-uncased-009 & - & 8 & 24 & 4e-5 & 43920\\
       BERT-base-uncased-010 & - & 16 & 24 & 4e-5 & 87840\\
       BERT-large-cased-wwm-003  & - & 2 & 4 & 3e-5 & 66040 \\
       BERT-large-uncased-wwm-eval-only  & SQuAD2 & - & - & - & - \\
\hline
       Roberta-base-uncased-eval-only  & SQuAD2 & - & - & - & - \\
       Roberta-base-uncased-001  & SQuAD2 & 8 & 24 & 4e-5 & 43944 \\
\hline
       Electra-base-eval-only  & SQuAD2 & - & - & - & - \\
       Electra-base-005  & SQuAD2 & 3 (20K) & 24 & 5e-5 & 2502 \\
       Electra-base-006  & SQuAD2 & 8 & 24 & 4e-5 & 43920 \\
\hline
       DistilBERT-base-unc-eval-only  & SQuAD2 & - & - & - & - \\
       DistilBERT-base-unc-003  & SQuAD2 & 5  & 24 & 5e-5 & 27450 \\
       DistilBERT-base-unc-004  & SQuAD2 & 5 & 32 & 5e-5 & 20500 \\
\hline
       Longformer-base-4096-eval-only  & SQuAD2 & - & - & - & - \\
       Longformer-base-4096-009  & SQuAD2 & 4 & 8 & 4e-5 & 65276 \\
\hline

    \end{tabular}
    \caption{Summary of our experiments, showing the models used, fine-tuning status, additional training epochs, batch size, and initial learning rate. The suffix \emph{eval-only} indicates evaluation runs posted weights.}
    \label{tab:BERT-models}
\end{table}

The models used for this evaluation are transformer-based, selected from the Huggingface Hub. We evaluated five BERT models that we fine-tuned on SQuAD v2. Using the published weights, we also evaluated various previously fine-tuned models. We then added a few epochs of fine-tuning and re-evaluated them. \autoref{tab:BERT-models} lists the resulting 16 experiments and the models involved. All experiments generated a list of the best ten predictions for analysis. For details about the models used, please refer to \autoref{sec:models}

\section{Analysis}
\input{31A_EM_F1_GR_Relationship}
\input{32A_GR_EM_Hist}
\input{33A_Clustering}
\input{34A_trainVals}

\input{45A_DirectPerformance}

%% file: 31A_EM_F1_GR_Relationship.tex
\subsection{Relationships among GR,  EM, and F1}

One of the concerns with introducing a new metric is avoiding measuring the same thing as other existing metrics. We consequently use the results of the validation runs from the experiments to visually demonstrate how GR and GRIM differ from EM and F1. One way to show these differences is by way of correlations. 

\begin{table*}[ht]
\begin{tabular}{|c|c|c|c|}
      \hline
      \includegraphics[width=0.23\textwidth]{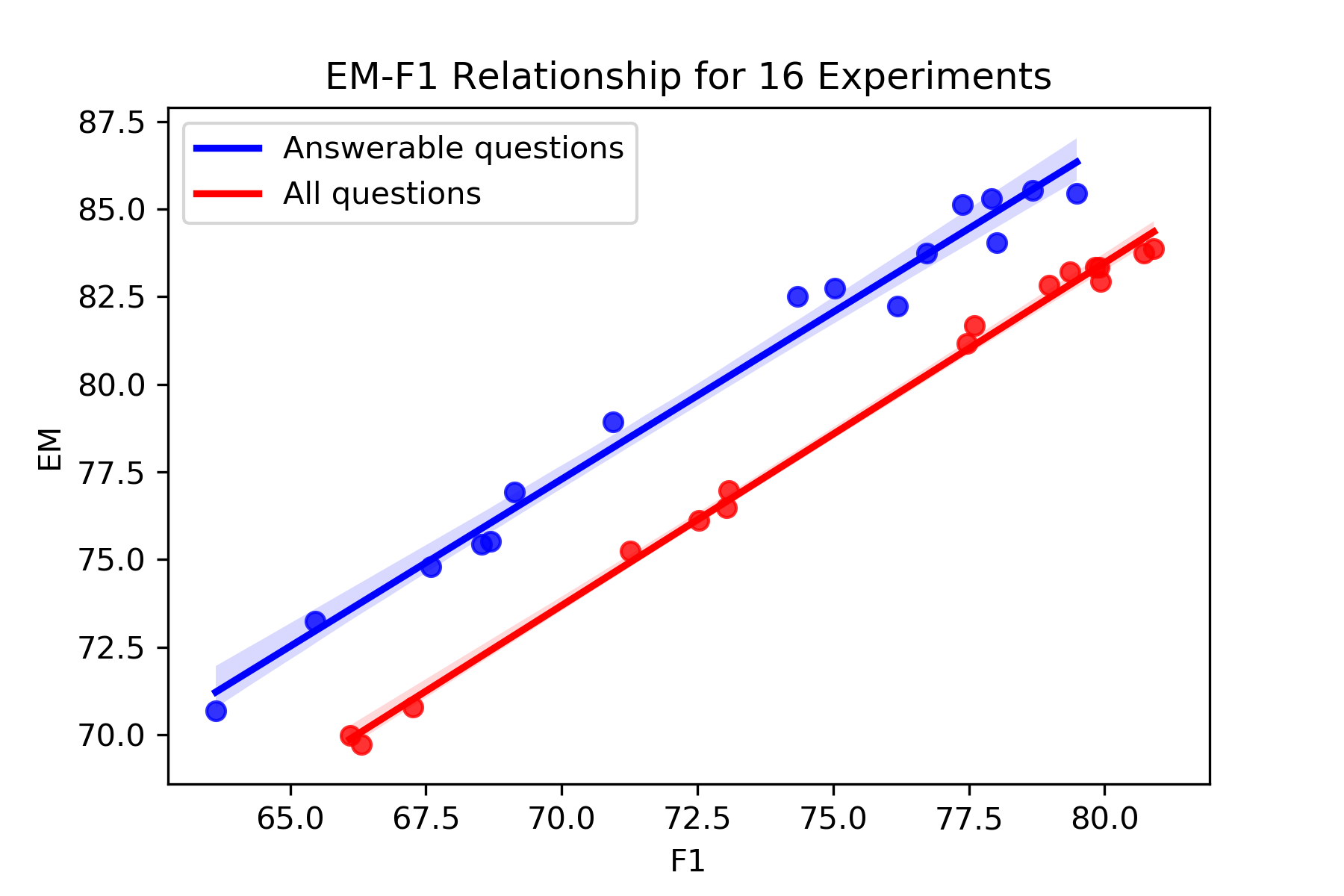} &
      \includegraphics[width=0.23\textwidth]{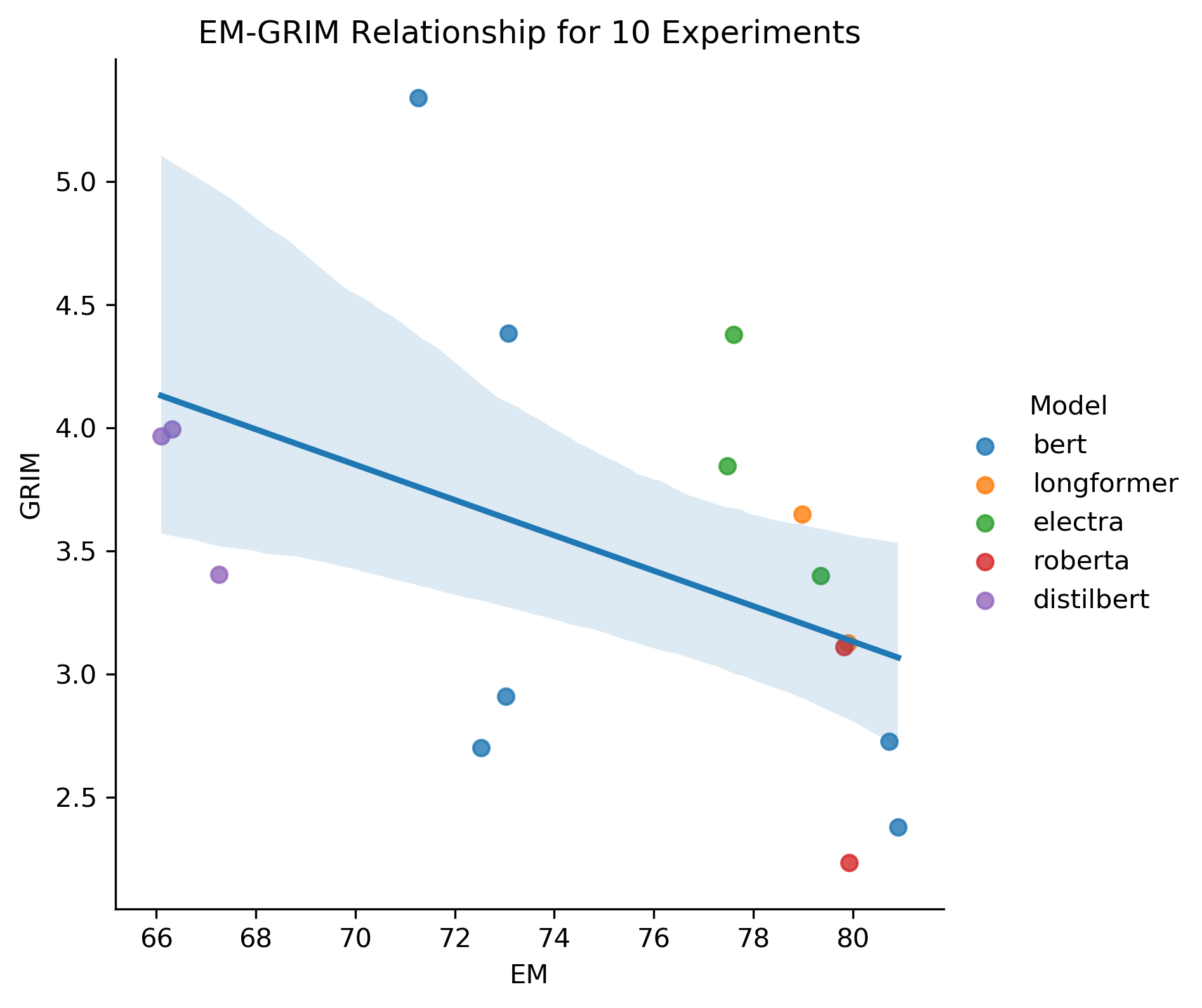} &
      \includegraphics[width=0.23\textwidth]{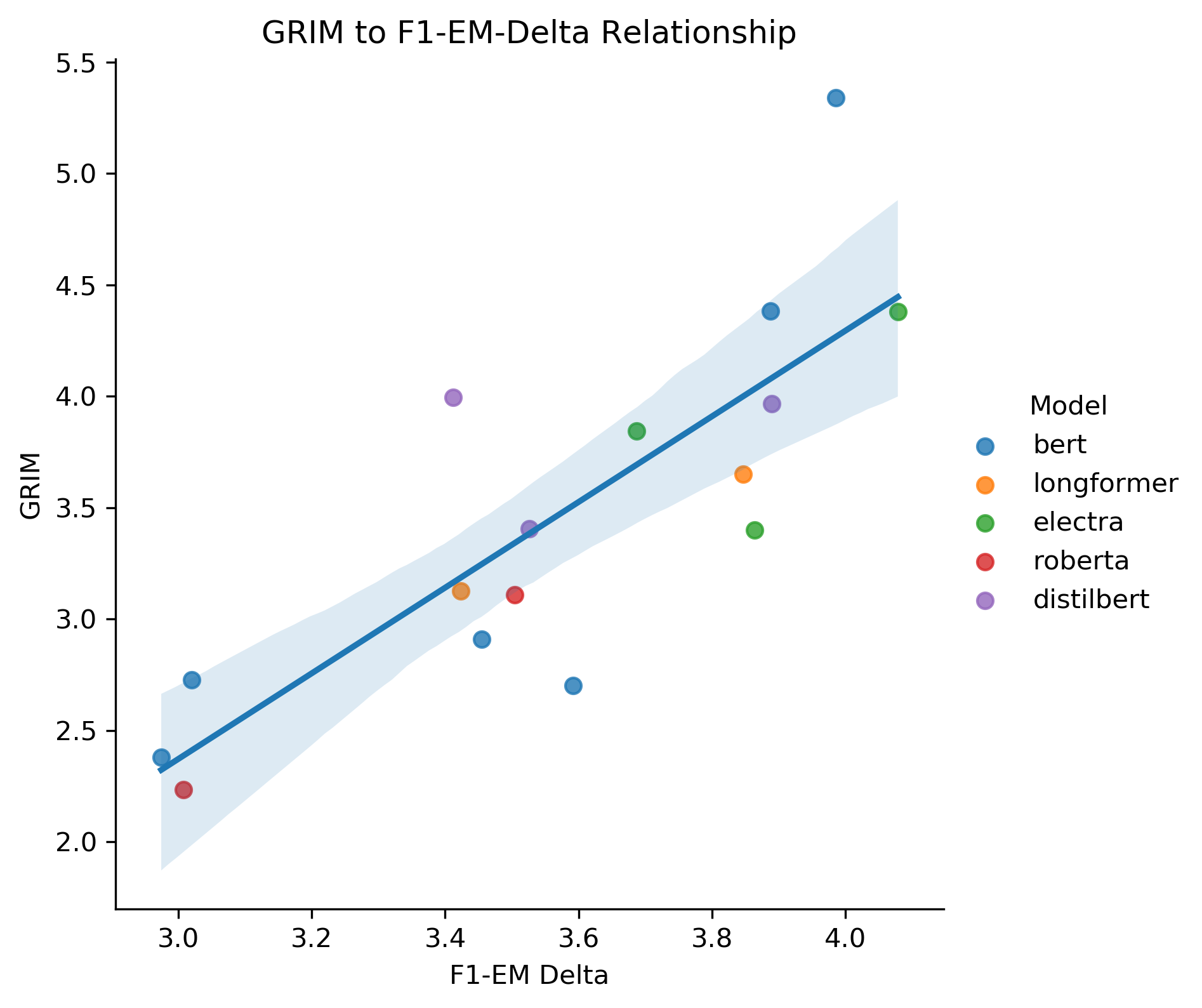} &
      \includegraphics[width=0.23\textwidth]{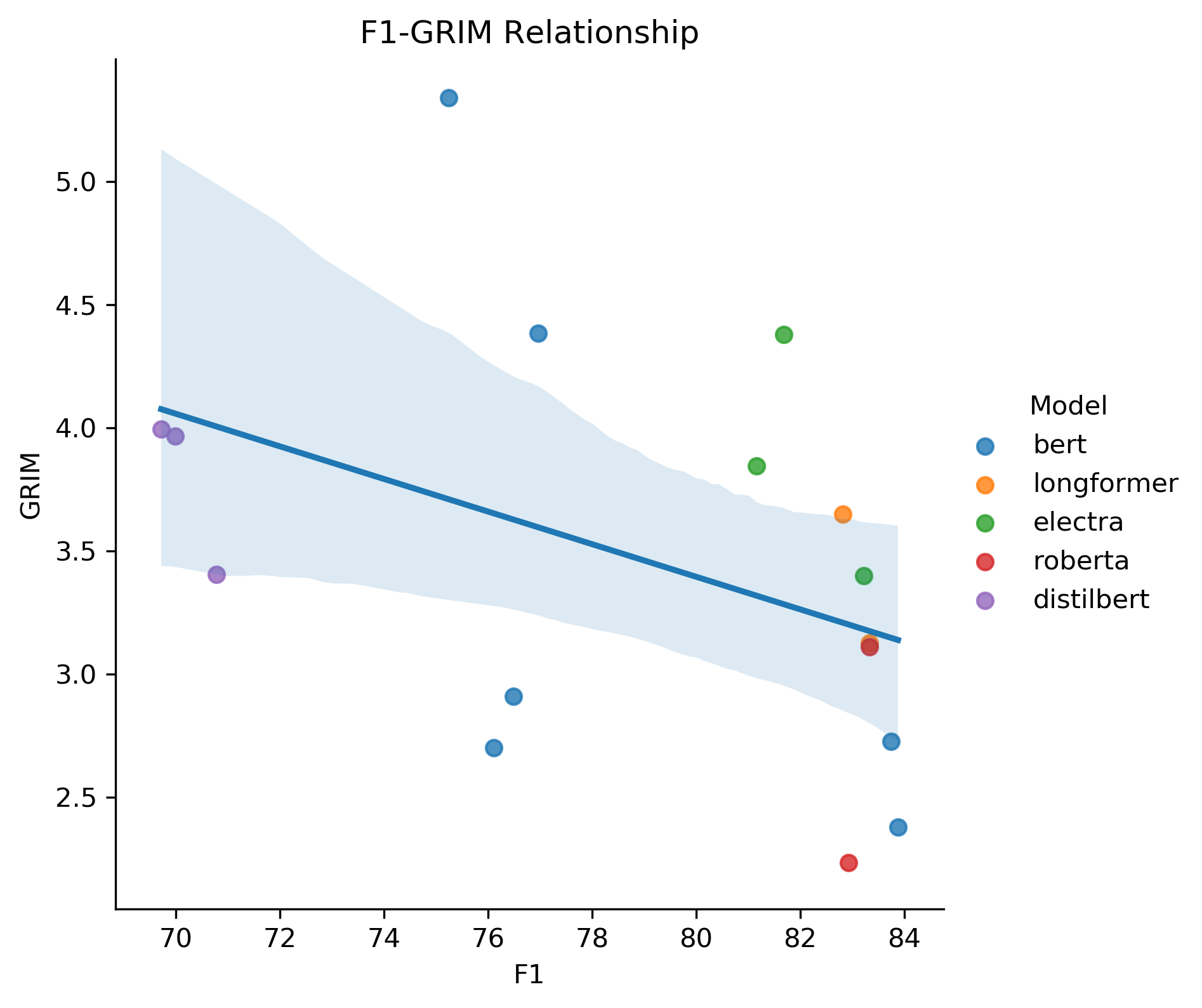} \\
      \hline
      
\end{tabular}
\caption{These plots show the independence between GRIM and EM or F1. The left graph demonstrates a high correlation between F1 and EM over 16 experiments on various transformer architecture models. The rest of the plots graphically show the low correlation between GRIM and EM, F1-EM, and F1, respectively. The dot colors indicate the models used.  }
\label{M:EM_F1_CORR}
\end{table*}

F1 can be viewed as a tie-breaker, being the same as EM for successful predictions and showing partial success for cases where the EM fails. However, the high correlation between F1 and EM  is  also because non-answers and monolectic answers always result in identical values for both metrics. The first plot on the left of \autoref{M:EM_F1_CORR} visually demonstrates the high EM-F1 correlation over all examples and the answerable examples separately. It shows the 16 experiments as points, the regression line, and the 95\% confidence region. 

We calculate the GRIM over secondary predictions only, which takes away a big (and the only) chunk of examples that would be perfectly correlated. The second plot of \autoref{M:EM_F1_CORR} shows the relationship between GRIM and EM, with the experiments as points, colored according to the transformer architecture behind each experiment. 

The remaining two graphs in \autoref{M:EM_F1_CORR} repeat similar GRIM plots vs. the F1 - EM difference and F1. The difference between F1 and EM is a more distilled measure of approximate success for failed predictions  derived from  standard metrics. Still, it fails to capture such proximity in the cases of non-answerable and monolectic answers.  

These visualizations are reasonably convincing that what we measure with GRIM differs from EM or F1. Also, the GR can be used to classify and analyze errors in secondary predictions. 

%% file: 32A_GR_EM_Hist.tex
\subsection{What GR and GRIM tell us about Models and Experiments}

We first examine the behavior of the new metrics at the model and experiment levels using results from the 16 experiments in  \autoref{tab:BERT-models}. For each experiment, we show the frequency distribution of examples by GR\footnote{The golden rank (GR), is the lowest ranking exact match to a correct answer} and, with dotted vertical lines, the GRIM as the interpolated median of GRs greater than zero.   To make this analysis more readable, we subdivided the experiments around a few themes, which we present below. All evaluations use the SqUAD-v2 validation dataset. 

We set forth an informal hypothesis we call the ``GRIM-performance hypothesis", stating that better-performing models tend to have lower GRIMs. However, our early analysis indicated that what is happening is more nuanced. So we introduce the notion of ``training maturity"\footnote{Not to be confused with MLOps operational maturity} as an informal measure of how early or late the best model emerges during training. We use the step (or epoch) that maximizes the F1 score to measure training maturity. Best models sometimes get established in only a couple of epochs, but more often, this occurs later in the training cycle. Our observations show evidence that all else being equal, experiments with comparable training maturity are more likely to uphold the GRIM-performance hypothesis, and higher training maturity makes this behavior more predictable.

\renewcommand{\arraystretch}{1.5}
\begin{table*}[ht]

\begin{tabular}{ccc}
      \rule{0pt}{75pt}
      \includegraphics[width=0.32\textwidth]{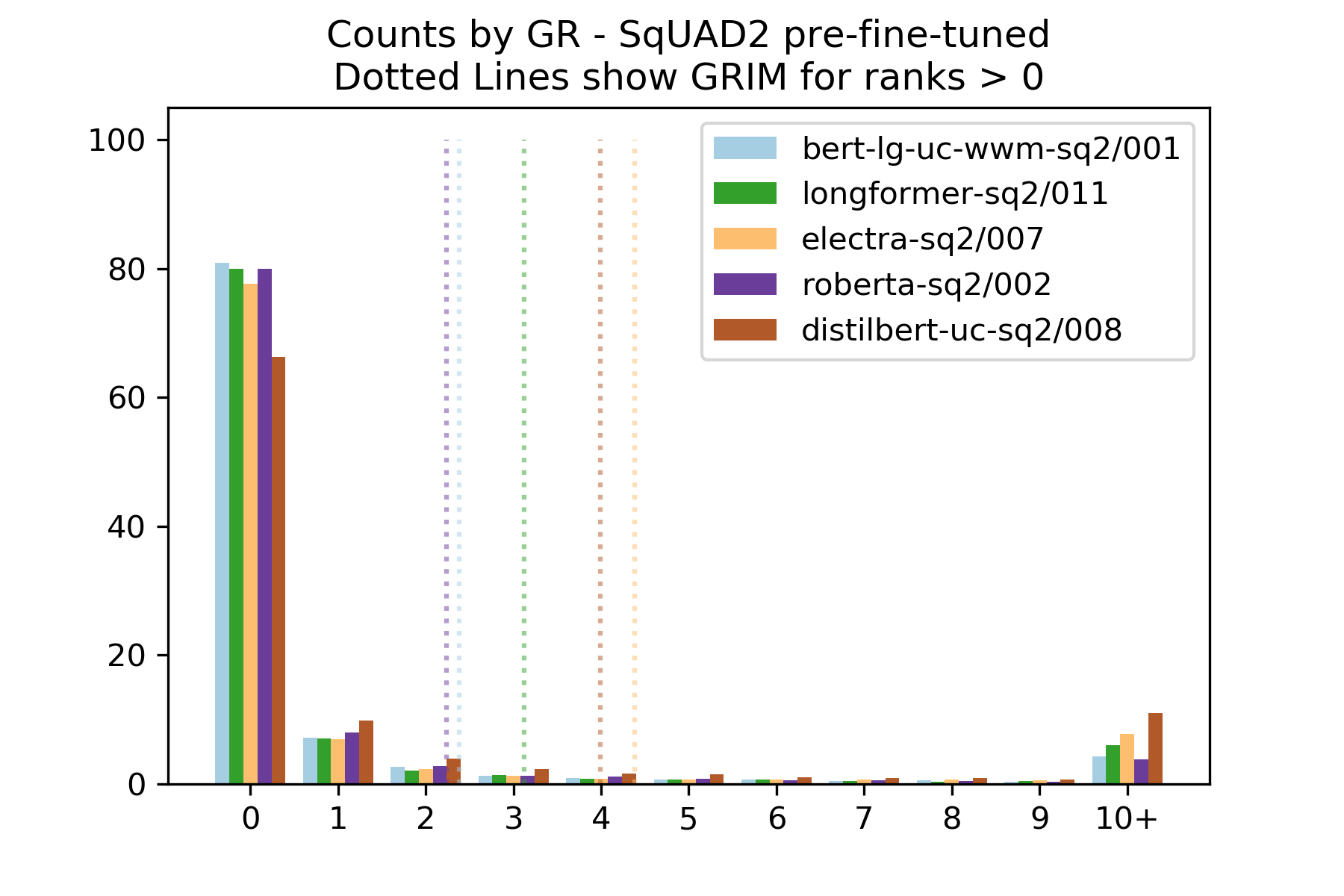} &
      \includegraphics[width=0.32\textwidth]{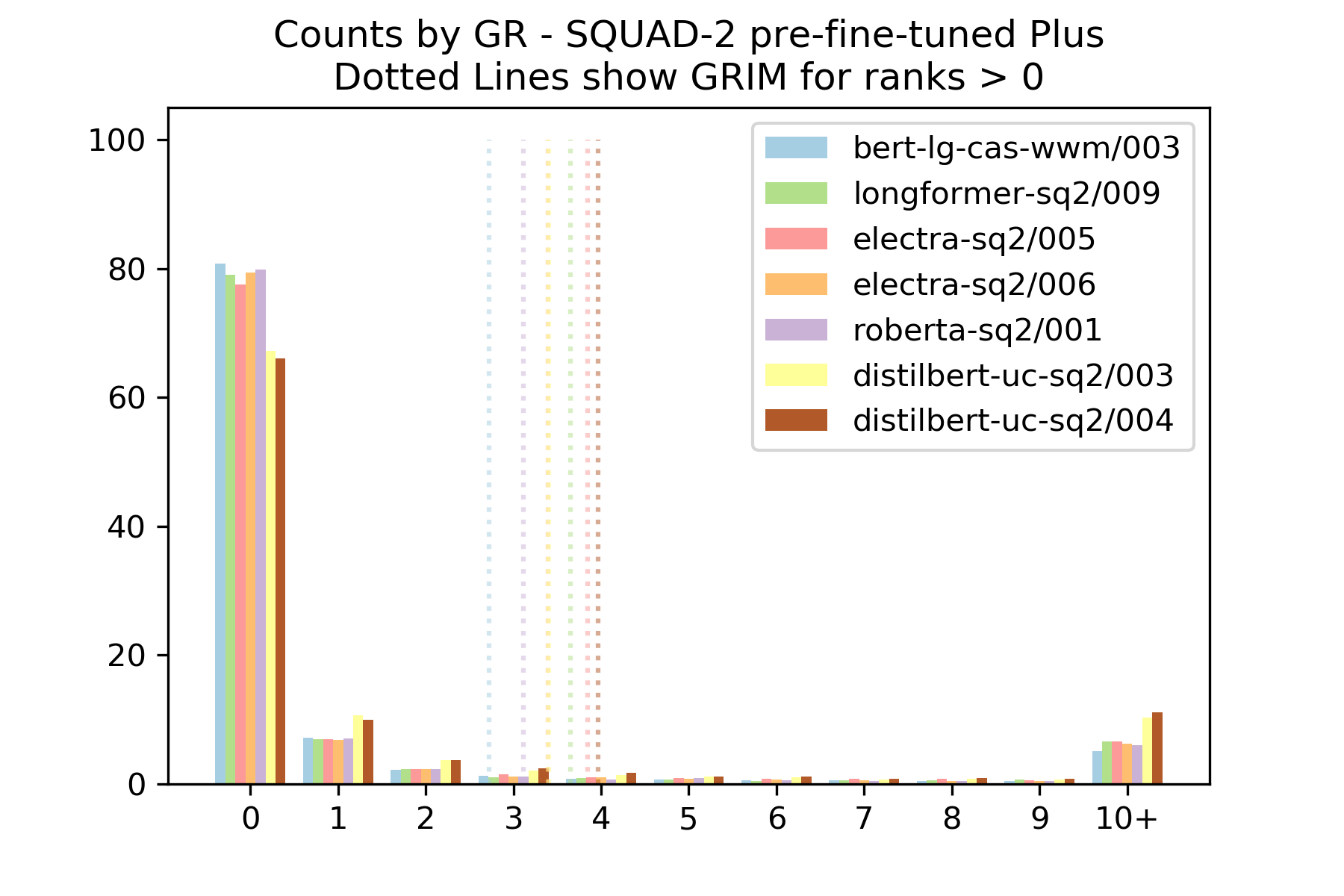} &
      \includegraphics[width=0.32\textwidth]{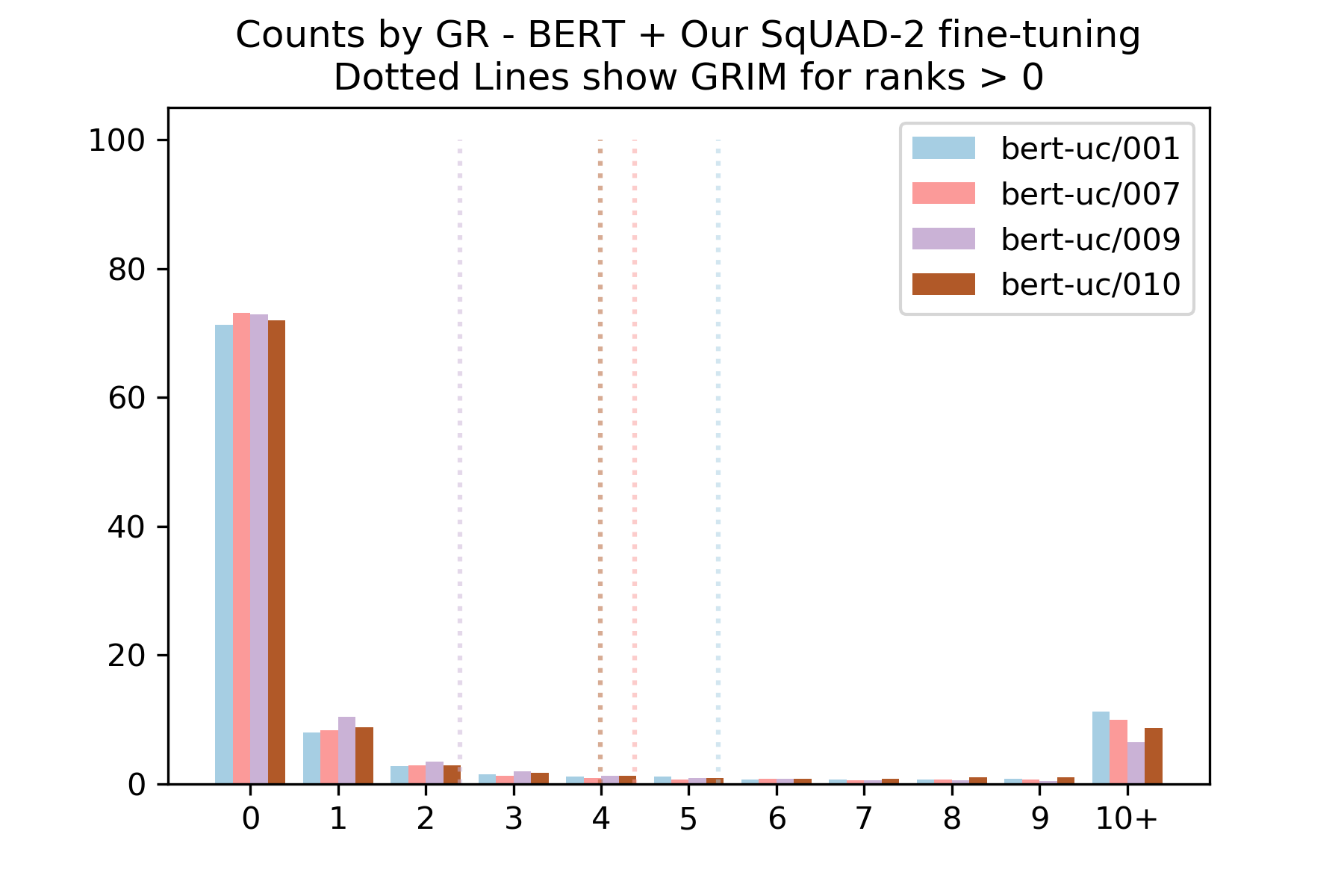} \\
      \rule{0pt}{75pt}\includegraphics[width=0.32\textwidth]{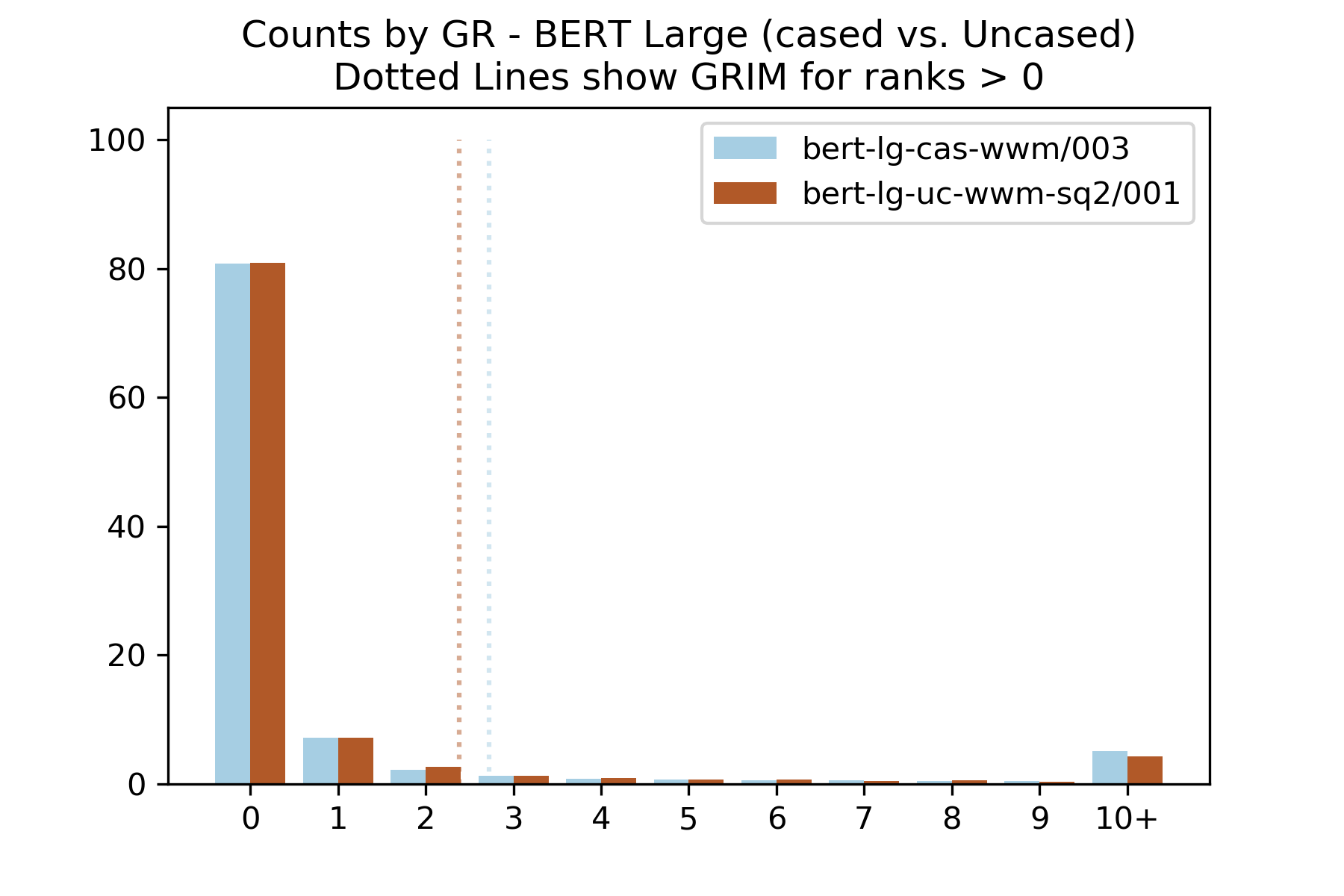} &
      \includegraphics[width=0.32\textwidth]{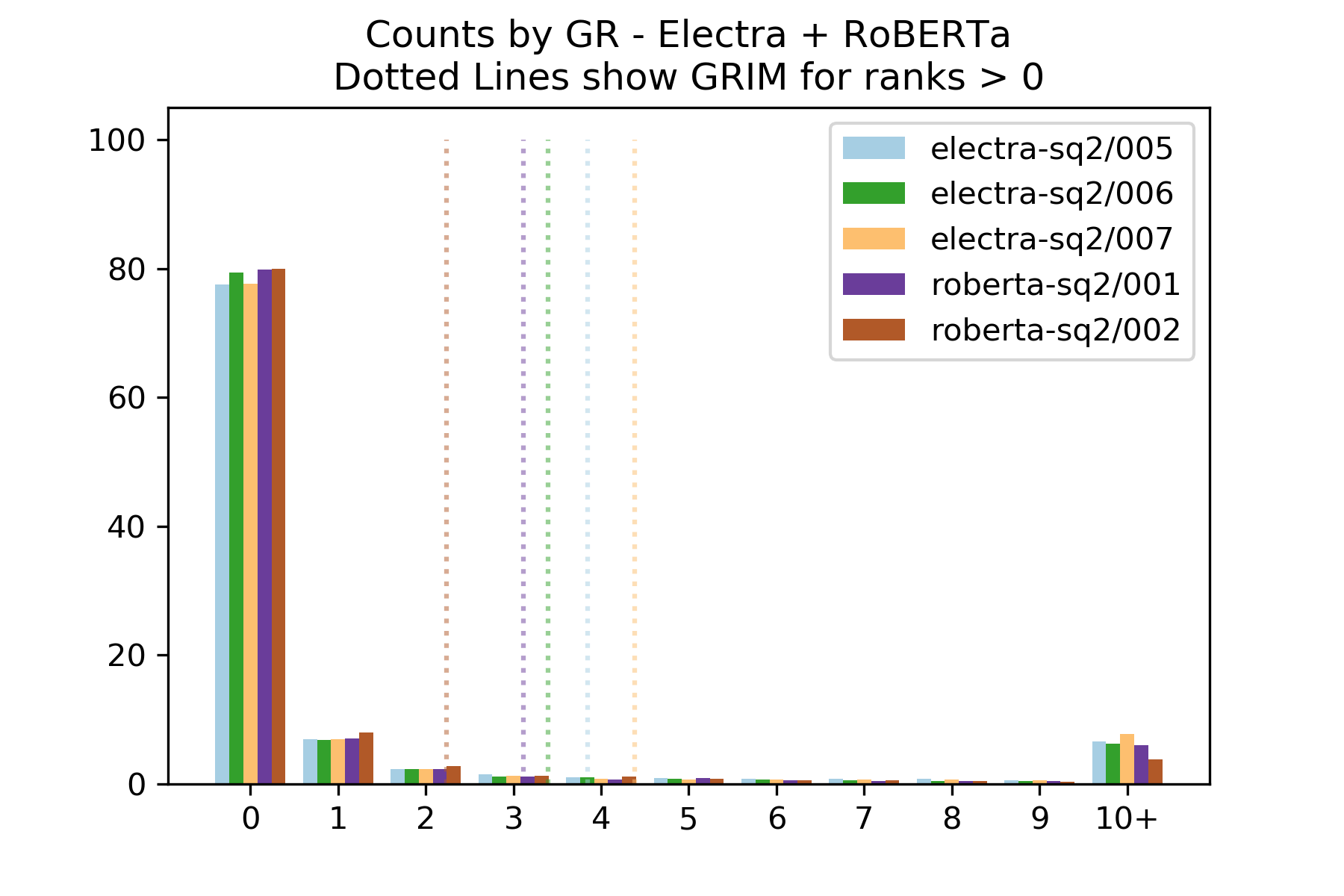} &
      \includegraphics[width=0.32\textwidth]{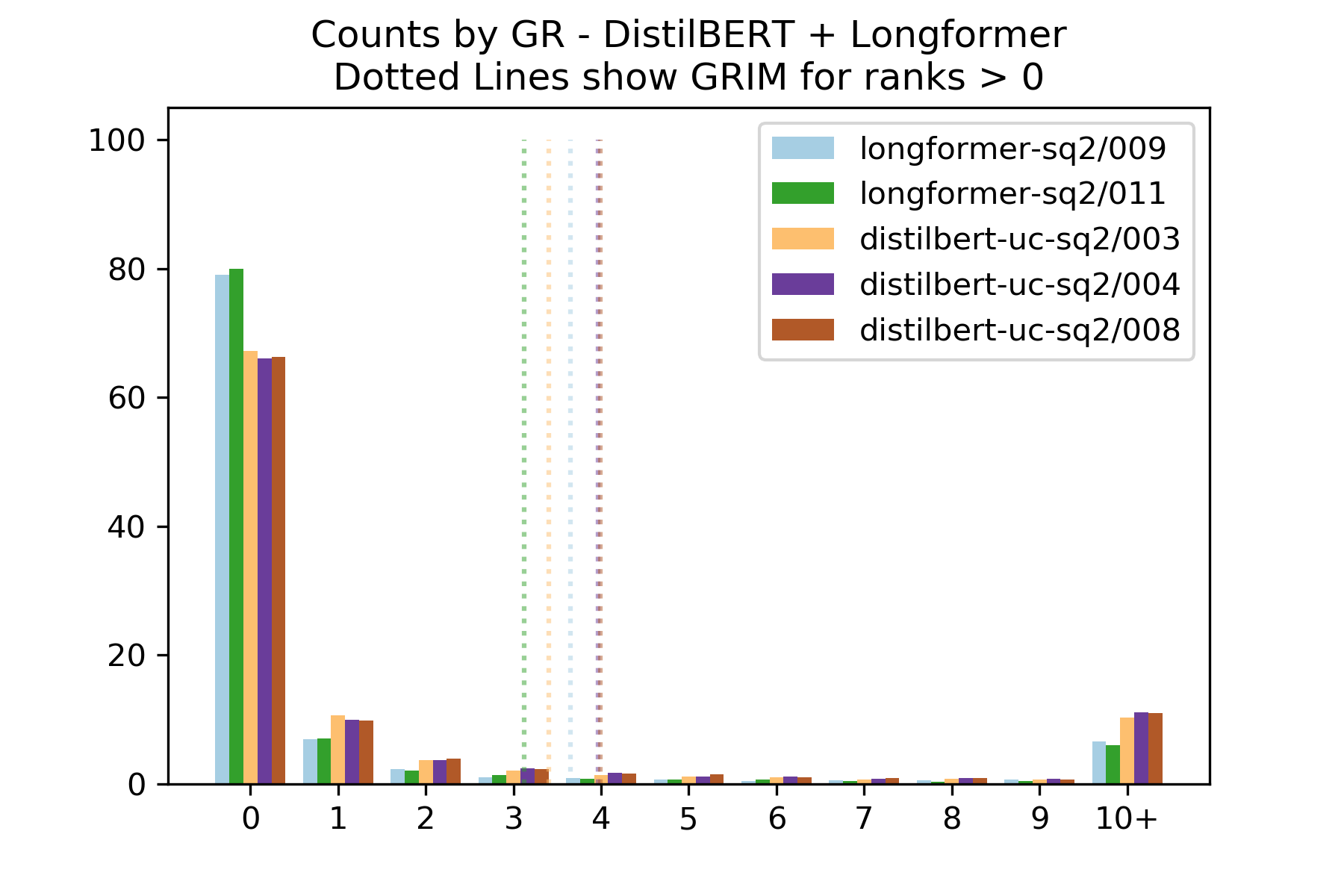} \\
\end{tabular}

\caption{Histograms showing collections of experiments by theme: The left-most bar shows the percentage of correctly answered examples, i.e., exact match. GRs 1-9 are shown over the next nine bars diminishing with increasing rank. The bar marked 10+ contains the aggregate count of all remaining examples with $GR \geq 10$. The dotted lines mark the GRIM for the respective experiment, which is calculated over secondary predictions only, i.e., where $GR>0$ }
\label{M:GR-EM-GRIM}
\end{table*}

Our rationalization is that many predictions are near-misses early in the training cycle, with their GR hovering around the low secondary ranks, pulling the GRIM lower. As the model improves, these low-hanging fruit predictions get absorbed into rank 0, and the remaining secondary GRs push the GRIM higher.  This behavior continues through training but is more volatile and profound early, where the effective learning rates are higher and later diminish towards a state of equilibrium. In \autoref{sec:GRIM-during-training}, we look at how GRIM, F1 and EM evolve during training, shedding light on this phenomenon. A word of caution: Fine-tuning is often compounded, done in multiple runs, often extending prior fine-tuning whose details may be unknown. In this analysis, we forego any attempt to normalize the measure of training maturity over the total length of training or batch size or other potentially relevant factors.
Consequently, our definition of training maturity is somewhat naive and purely qualitative. Attempting a more robust method requires extensive study with more models and other classification tasks. It should also consider factors such as training length, batch size, overfitting, and effects such as the model's generalization capability.

\subsubsection{Pre-Fine-Tuned Models}

The top-left histogram in \autoref{M:GR-EM-GRIM} shows results from evaluation runs on the five already fine-tuned models we obtained from the Hugging Face Hub. These five tests used the published weights with no additional fine-tuning. Scores are shown in the top section of \autoref{tab:BERT-pretrained-1}. A few notable observations are:

\begin{enumerate}
    \item The best two performing models, BERT-large and  RoBERTa, have low GRIM. Their secondary GR distribution leans to the left, with fewer examples in the 10+ rank than the rest of the models. They uphold the GRIM-performance hypothesis except for Electra. 
    \item The worst performing model is DistilBERT, with a high GRIM and a significant spike at rank 10+. It upholds the GRIM-performance hypothesis except for Electra.  
    \item With performance similar to RoBERTa's, Longformer  has more examples in the 10+ rank than either Roberta or BERT-large, which raises its GRIM midway between Roberta's and DistilBERT's. It upholds the GRIM-performance hypothesis except for Electra. 
    \item Electra, although relatively well-performing, exhibits contrarian behavior relative to the GRIM-performance hypothesis, with secondary GR distribution leaning towards higher ranks and GRIM higher than the worse-performing DistilBERT. 
\end{enumerate}

\begin{table}
    \centering
    \begin{tabular}{l|c|r|r|r|r}
    \hline
     Model/Experiment & Total & EM    & Best-F1& F1     & GRIM \\
     Description      & Steps & Score & Step   &  Score & Score\\
\hline
       BERT-lg-uc-wwm-sq2-001  & - & 80.91 & - & 83.88 & 2.38  \\
       Longformer-sq2-011      & - & 79.90 & - & 83.33 & 3.13  \\
       Electra-sq2-007         & - & 77.61 & - & 81.68 & 4.38  \\
       Roberta-sq2-002         & - & 79.92 & - & 82.93 & 2.23 \\
       DistilBERT-uc-sq2/008   & - & 66.31 & - & 69.72 & 3.99  \\
\hline
       BERT-lg-cas-wwm-003     & 66K & 80.73 & 61.5K & 83.75 & 2.73 \\
       Longformer-sq2-009      & 65K & 78.98  & 55K & 82.82 & 3.65 \\
       Electra-sq2-005         & 2.5K& 77.47  & 2.5K & 81.16 & 3.84 \\
       Electra-sq2-006         & 44K & 79.36  & 44K & 83.22& 3.40 \\
       Roberta-sq2-001         & 44K & 79.83  & 27.5K & 83.33 & 3.11 \\
       DistilBERT-uc-sq2-003   & 27K & 67.27  & 21K & 70.79 & 3.41 \\
       DistilBERT-uc-sq2-004   & 21K & 66.1   & 20.5K & 69.99 & 3.97 \\
\hline

    \end{tabular}
    \caption{ Results from comparing the first two graphs with pre-trained models evaluated with the published weights in the top section and the same models after we added more fine-tuning in the lower section. The table shows EM, F1, GRIM, total steps, and training maturity as the step   where the best model, based on the F1 score, was achieved.}
    \label{tab:BERT-pretrained-1}
\end{table}

\subsubsection{Same Models with Additional Fine-Tuning}

The middle-top graph in \autoref{M:GR-EM-GRIM}  shows the same fine-tuned models after adding more fine-tuning runs (see table \ref{tab:BERT-models} for details). The additional fine-tuning causes a few  subtle changes worth noting:

\begin{enumerate}
    \item The ``spread" of the GRIMs between best and worst-performing has narrowed.
    \item The DistilBERT-003 experiment in yellow has substantially improved and upholds the GRIM-performance hypothesis in all cases except when compared to Longformer 009 and Electra 005.
    \item Electra's new experiments showed significantly lower GRIM than the experiments with the published weights. Notice that the more successful Electra experiment (006) has  the lowest of the three Electra GRIMs, supporting the GRIM-performance hypothesis. 
    \item For BERT-Large (003), Roberta (001) and Longformer have slightly lower EM scores than in the earlier graph, and the GRIM moved higher, upholding the GRIM-performance hypothesis within the same architecture.
\end{enumerate}

\subsubsection{BERT-Uncased}

The top-right graph  in \autoref{M:GR-EM-GRIM}  shows four fine-tuning experiments we performed\footnote{the second version of this document replaced experiments 009 and 010 with new ones. The original experiments were interrupted and restarted, which caused minor errors in the processing of the results.} on the pre-trained BERT-base-uncased model, with no prior fine-tuning. Results are shown in \autoref{tab:BERT-models2}. There are three distinct training maturity horizons among the four examples, 16.5K, 30.5K, and 56K. The first three experiments are consistent with the GRIM-performance hypothesis in conjunction with training maturity conditioning. The last experiment, BERT-010, with the latest maturity, has a lower GRIM than BERT-007, which has the best score. This pattern contradicts the originally stated GRIM-performance hypothesis and  maturity conditioning assumption.

\begin{table}
    \centering
    \begin{tabular}{l|c|r|r|r|r}
    \hline
     Model/Experiment & Total & EM & Best-F1& F1 & GRIM \\
     Description & Steps & Score & Step &  Score & Score\\
\hline

       BERT-base-uncased-001 & 44K & 71.3 & 30.5K & 75.25 & 5.3\\
       BERT-base-uncased-007 & 44K & 73.1 & 30.5K & 76.97 & 4.4 \\
       BERT-base-uncased-009 & 44K & 72.92 & 16.5K & 76.49 & 2.4\\
       BERT-base-uncased-010 & 88K & 71.94 & 56K & 75.78 & 4.0\\
\hline

    \end{tabular}
    \caption{BERT-base experiments, showing EM, F1, GRIM, and the optimization step of the best F1 score. }
    \label{tab:BERT-models2}
\end{table}

\subsubsection{BERT-Large Experiments } 
\label{sec:Bert-Large-Exp}
The bottom-left graph in \autoref{M:GR-EM-GRIM} shows two BERT-large experiments, the first (cased), evaluated after we applied two epochs of fine-tuning, and the second  (uncased), fine-tuned by its publishers. \autoref{tab:BERT-pretrained-1}  shows their low GRIM scores and high performance, which is consistent with the GRIM-performance hypothesis. In this case, the low GRIM resonates with the high performance, but their relatively low training maturity may also cause the low GRIM values. These larger models, being more efficient learners and  costlier to train, are typically fine-tuned over fewer epochs. The step at which the best model occurs is deceivingly high, but when normalized for the smaller batch size, it becomes $\frac{24}{4}=6$ times lower compared to the base models, i.e., barely over 10K of equivalent steps. In \autoref{sec:GRIM-during-training}, we show an experiment where we fine-tuned this BERT-large-cased model for six epochs to better understand the relationship between performance, training maturity, and GRIM. 

\subsubsection{Other Consolidated Comparisons by Model Architecture} 
\label{sec:robeta-Exp}

The lower middle graph in \autoref{M:GR-EM-GRIM} consolidates  
all RoBERTa and Electra experiments. Similarly, the lower right graph does the same for 
all  Longformer and DistilBERT experiments. These experiments have been  discussed earlier, with their scores and training maturities included  in \autoref{tab:BERT-pretrained-1} 

%% file: 33A_Clustering.tex
\subsection{What GR Distributions tell us about Individual Examples}

Example-level GR distributions over experiments are helpful for error analysis or example difficulty assessment. The most straightforward examples have zero mean and zero standard deviation. The hardest ones, perhaps those with erroneous annotations, have a mean of 10 and zero standard deviation. For the rest, the mean and standard deviation are telling in characterizing them.

The four graphs in table \ref{M:CLUSTERS} show each example as a point with the x-axis denoting the GR mean over the 16 experiments 
and the y-axis representing the standard deviation. As discussed in section \ref{sec:30A}, our ranking scale has an artificial catch-all maximum of 10 for all GRs higher than 9. So, the right half of the graph is distorted, with y-values artificially vanishing at ten and appearing symmetric to the left half, making it take a dome-like shape. 
If we used the actual rank values, which can be in the hundreds or thousands, the right side would extend into a semi-parabolic open-ended shape with sparser and sparser density towards the right. 

To better visualize the constraints that shape the graph, consider that the only way for the mean to be 0 (or 10) is for all 16 experiments to produce GR 0 (or 10) and thus to have no standard deviation other than zero. Also, the highest standard deviation that an example can reach: 5, is only achieved when the GR of eight experiments is 0 and the other eight 10. The higher an example appears, the more dispersed and polarized the GR scores evolve.   \autoref{sec:A30} presents a few highly polarized examples in more detail.

With this in mind, let us look at our analysis.  

\renewcommand{\arraystretch}{1.5}

\noindent
\begin{table*}[ht]
\begin{tabular}{cc}
      \rule{0pt}{75pt}
        \includegraphics[width=0.5\textwidth]{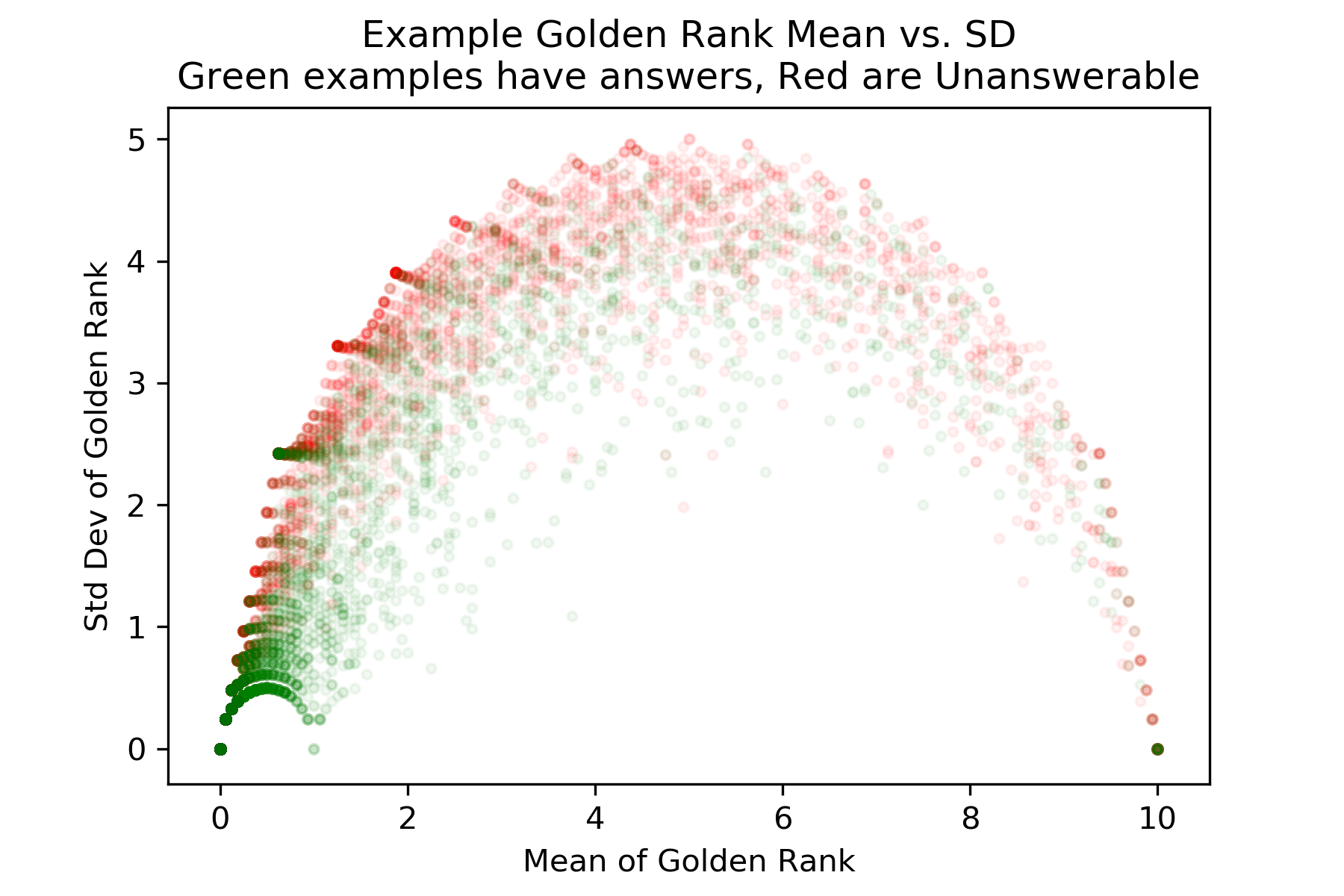} &
        \includegraphics[width=0.5\textwidth]{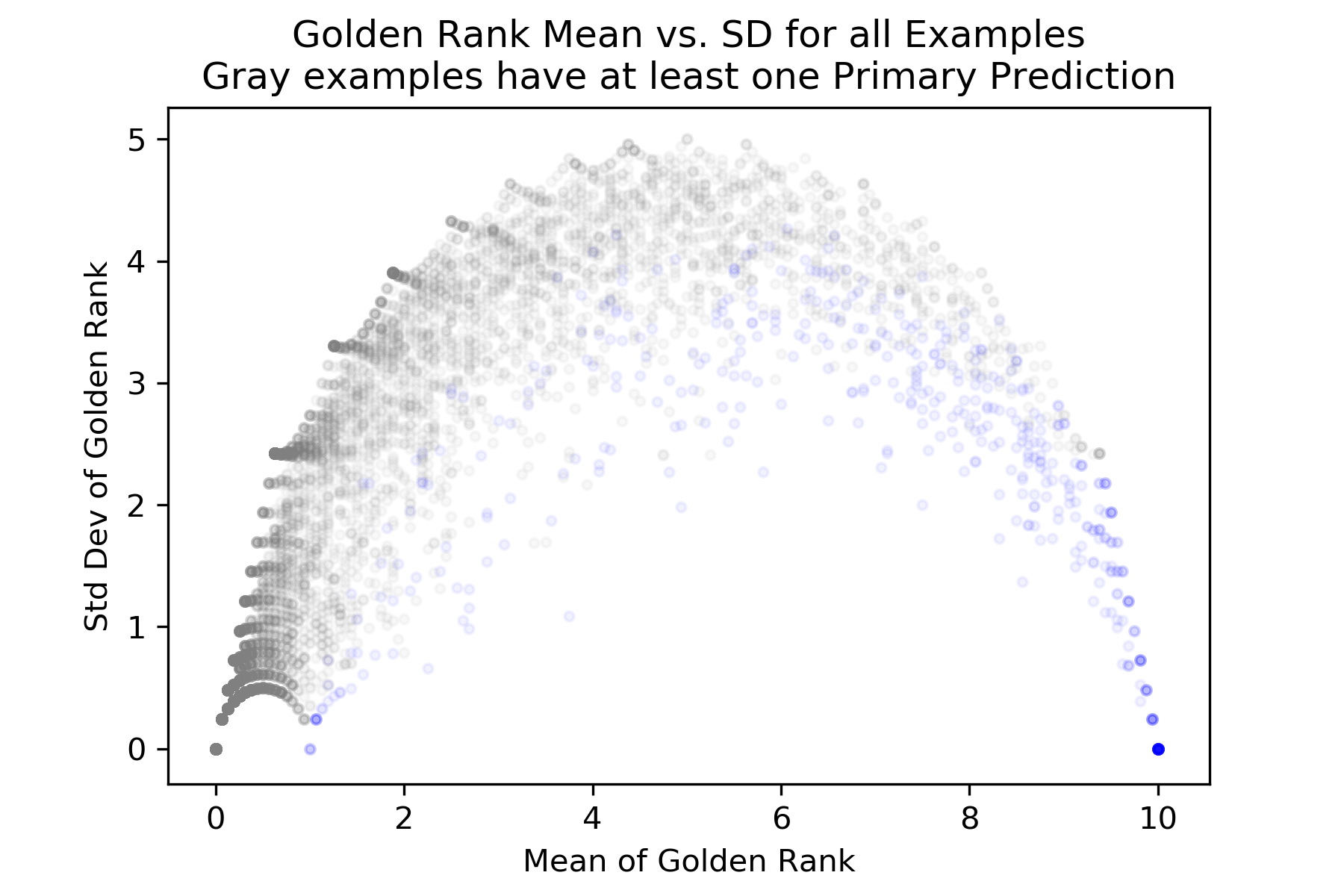} \\
      \includegraphics[width=0.5\textwidth]{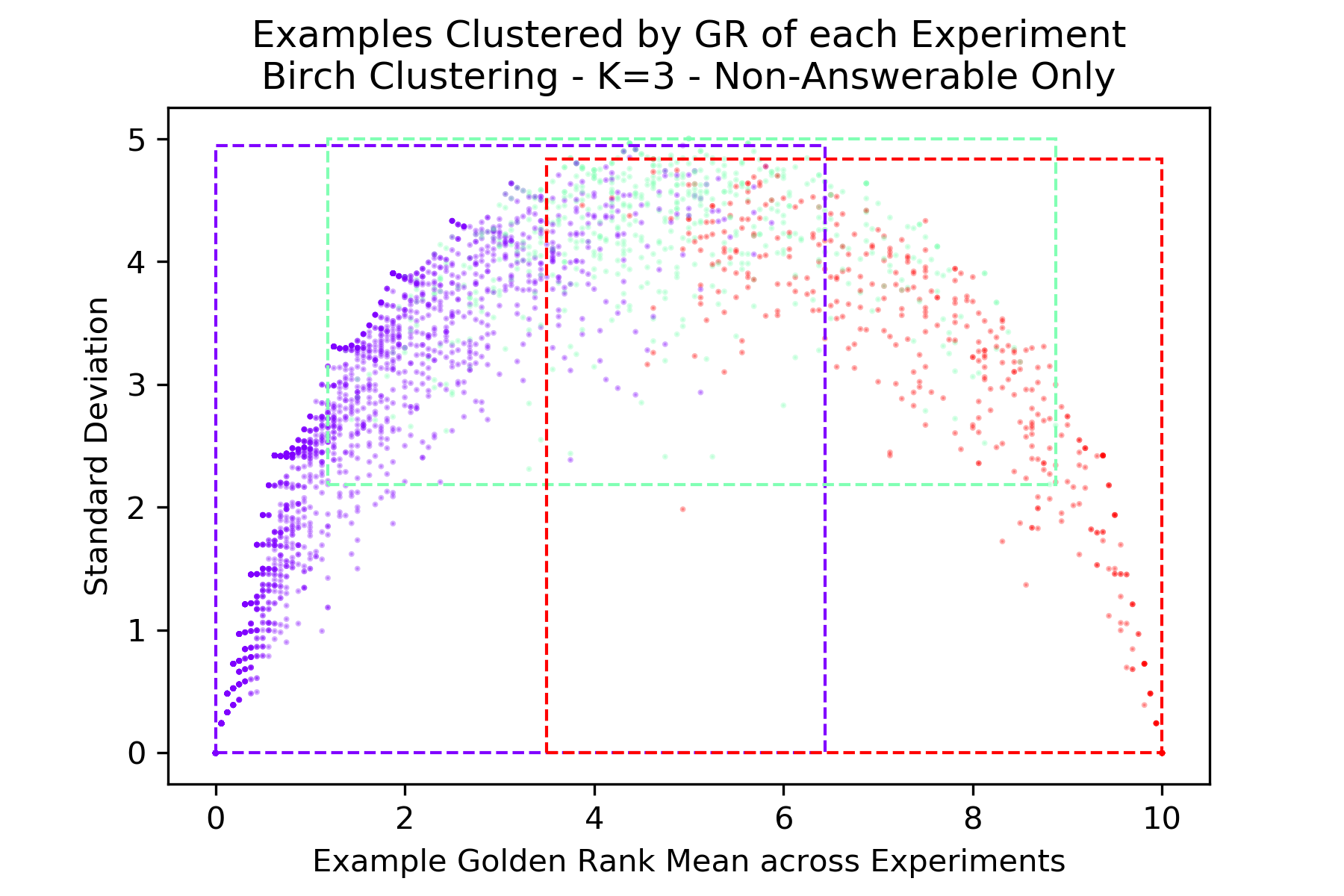} &
      \includegraphics[width=0.5\textwidth]{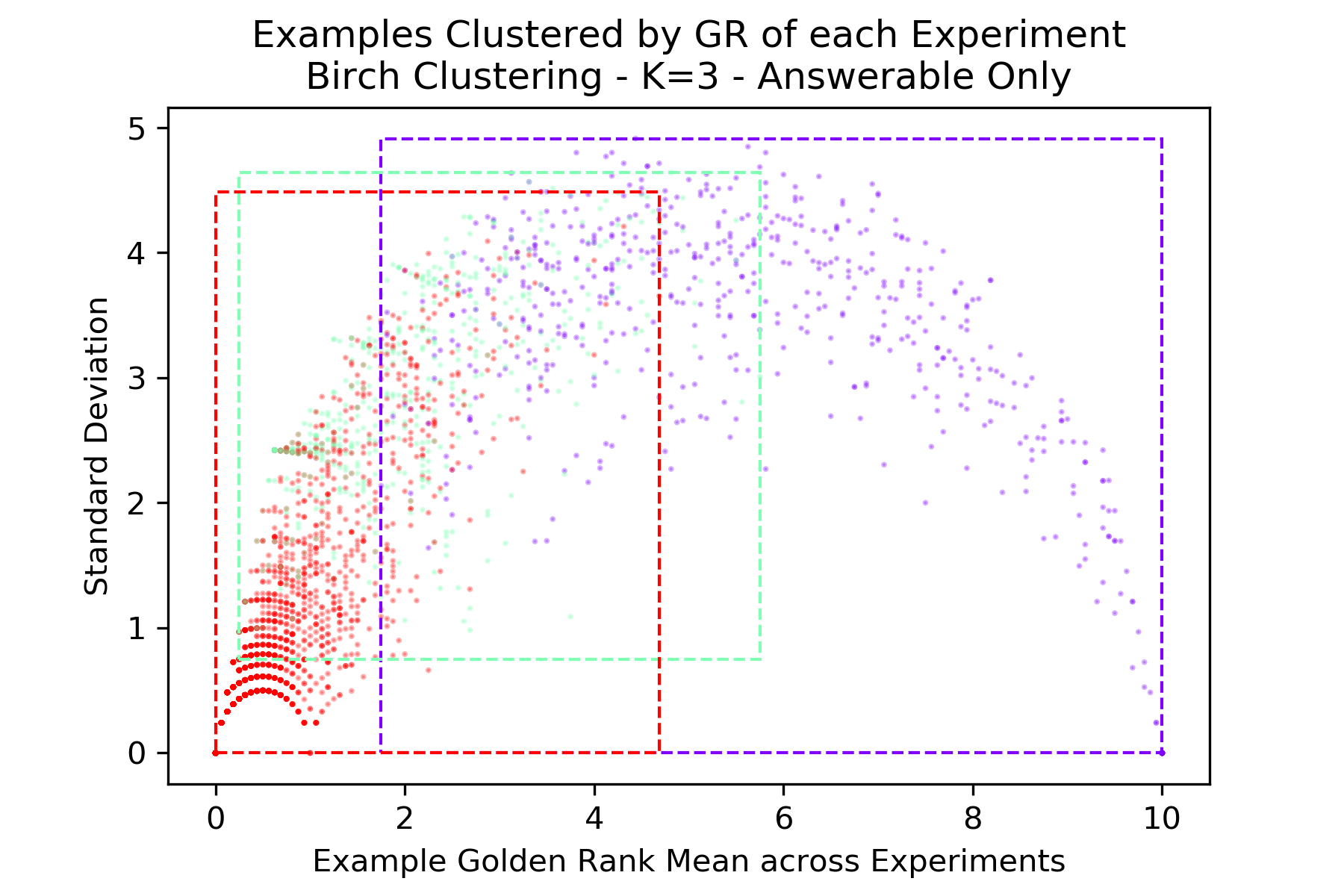} \\
\end{tabular}
\caption{Each dot represents a QA example, placed according to the mean and standard deviation of its Golden Ranks from the 16 experiments. Top-left: All examples, answerable in green vs. non-answerable in red. Top-right: All examples, distinguishing those with at least one model predicting them correctly vs. those never predicted  correctly. Bottom: All examples, colors denote clusters  derived using the Birch agglomerative clustering algorithm (with K=3) on the 16 experiment GR values for non-answerable  and answerable questions separately}
\label{M:CLUSTERS}
\end{table*}

\subsubsection{Overview of All Experiments}

The top left diagram in table \ref{M:CLUSTERS} shows an overview of all examples. They are color-coded by whether they are answerable or not according to the golden answers of the dataset\footnote{The official SQuAD-v2 dataset uses a flag named ``is impossible" to denote those questions that cannot be answered from the context. We use the term non-answerable or unanswerable in this paper.}. The green represents answerable examples and the red non-answerable ones. There are 11,876 examples processed, from which 5945 are non-answerable and 5928 answerable\footnote{The dataset contains 5945 answerable examples, but 17 were discarded because they failed during processing}. The point at (0,0) represents all 4654 (39.20\%) examples that scored GR=0 for all 16 experiments. From those, 2538 (21.38\%) are unanswerable, and 2116 (17.82\%) are answerable. Point (10,0) represents 44  examples that scored $GR > 9$ for all experiments, 24 non-answerable and 20 answerable. They denote either challenging or perhaps wrongly annotated examples. The only other point with zero standard deviation is (1,0), which accounts for four answerable examples that consistently  scored 1 in all experiments, discussed in more detail in \autoref{sec:A21}. 

Non-answerable examples tend to have higher standard deviation and  to spread more towards higher ranks. The upper center section contains very polarized GR scores. Answerable examples concentrate on lower ranks and lower standard deviation, which alludes to models more consistently predicting them correctly. 

The top right graph shows a different slice of the examples. Using the EM criterion, 11397 examples, shown in black, have been correctly predicted $(GR=0)$ by at least one experiment, while the remaining 476 (4.0\%) were never predicted correctly. Considering the combination of mean and standard deviation, we can refine this categorization and  construct a more nuanced spectrum of difficulty to prioritize example inspections more efficiently and systematically. These inspections can serve the purposes of error analysis, difficulty assessment, or hunting down incorrect answers and annotations.

\subsubsection{Clustering of Experiments}

We use the GR mean and standard deviation to partition examples into clusters. We do this separately for the non-answerable and the answerable ones. 
The bottom-left figure in table \ref{M:CLUSTERS} contains only non-answerable examples using the same dot pattern, whereas the coloring denotes the clusters. Having tried various algorithms and numbers of groups $K$, we here show the results of the BIRCH hierarchical clustering algorithm \cite{10.1145/233269.233324} with $K=3$.  

\begin{table*}[ht]
\centering
\begin{tabular}{lrrrrr}
\toprule
Cluster &  Counts &  from GR &  to GR &  from Std &    to Std \\
\midrule
     All Correct &    2538 &     0.0000 &   0.0000 & 0.000000 & 0.000000 \\
  Mostly Correct &    2420 &     0.0000 &   6.4375 & 0.000000 & 4.943035 \\
       Polarized &     583 &     1.1875 &   8.8750 & 2.185714 & 5.000000 \\
       Challenges &     404 &     3.5000 &  10.0000 & 0.000000 & 4.833154 \\
\bottomrule
\end{tabular}
\end{table*}

More than half the examples of the purple cluster are at the point (0,0),  as the first line of the table above shows, capturing all correctly predicted un-answerable examples. The rest of the cluster exhibits a low average GR score which we labeled as ``Mostly Correct". The  ``Polarized" group includes examples with means around the middle of the range and high standard deviation.  The third cluster in red contains examples with GR-mean scores in the higher ranks that appear to pose a  challenge to most experiments.  

The bottom-right graph repeats the same exercise for answerable examples with cluster details as shown below:
\begin{table*}[ht]
\centering
\begin{tabular}{lrrrrr}
\toprule
Cluster &  Counts &  from GR &  to GR &  from Std &    to Std \\
\midrule
     All Correct &    2116 &       0.00 &   0.0000 &     0.00 & 0.000000 \\
  Mostly Correct &    2715 &       0.00 &   4.6875 &     0.00 & 4.485654 \\
       Polarized &     541 &       0.25 &   5.7500 &     0.75 & 4.639757 \\
       Challenges &     556 &       1.75 &  10.0000 &     0.00 & 4.911323 \\
\bottomrule
\end{tabular}
\end{table*}

The first (red) cluster is, again,  divided into two parts: the ``All Correct" with examples correctly predicted by all experiments at coordinate (0,0), and the rest of those that belong to the lower left region labeled as ``Mostly Correct". Notice that this cluster covers a smaller area with a similar count as the corresponding cluster in the unanswerable plot for the answerable examples. Also, the polarized group is narrower and less polarized than the unanswerable equivalent.  

SQUAD-2 introduced unanswerable questions to raise the level of difficulty \cite{rajpurkar2018know}, so the differences between the two graphs above are not a surprise.

%% file: 34A_trainVals.tex
\subsection{GR/GRIM Movement During Training}
\label{sec:GRIM-during-training}
The next aspect of our analysis  is understanding how the GRIM moves during the training cycle relative to the EM and F1 scores. We compare EM, F1, and GRIM from all validations performed during two different training experiments. Their plots are  in table \ref{M:EM-GRIM}. These experiments were on pre-trained but not fine-tuned models. The graph on the left uses a  RoBERTa-base model trained over 16 epochs with batch size 24. The one on the right uses a BERT-large model for six epochs and batch size 5. 
Notice that the last evaluation is always a repetition of the evaluation using the best model encountered during the training run.

Observing the movements relative to the three metrics in these graphs, we can observe three characteristics of the GRIM:

\begin{itemize}
    \item A continuous oscillation is combined with an upward trend from lower ranks (2 or 3) to a higher rank, around 5. 
    \item The oscillation magnitude has a pattern of contracting and expanding with a periodicity that differs for the two models, i.e., every few validations for Roberta-base, but more sustained longer intervals in BERT-large. This oscillation diminishes as training matures.
    \item The GRIM movement is in a direction  opposite to that of the  EM and F1.
\end{itemize}

Notice that the left y-scale is for the EM and F1 scores, while the right y-scale depicts the rank for the GRIM.

\renewcommand{\arraystretch}{1.5}

\noindent
\begin{table*}[ht]
\begin{tabular}{cc}
      \rule{0pt}{75pt}
      \includegraphics[width=0.5\textwidth]{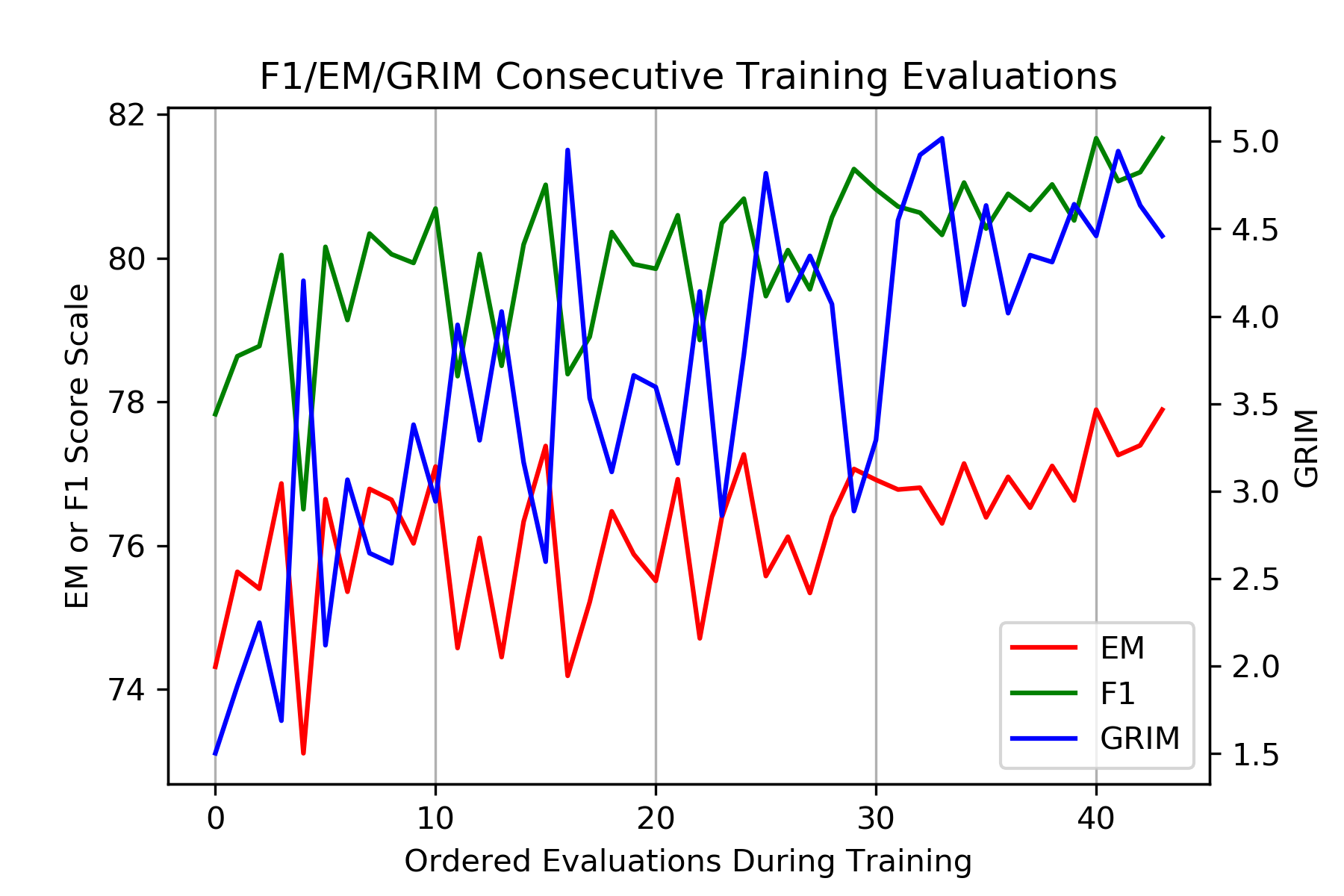} &
      \includegraphics[width=0.5\textwidth]{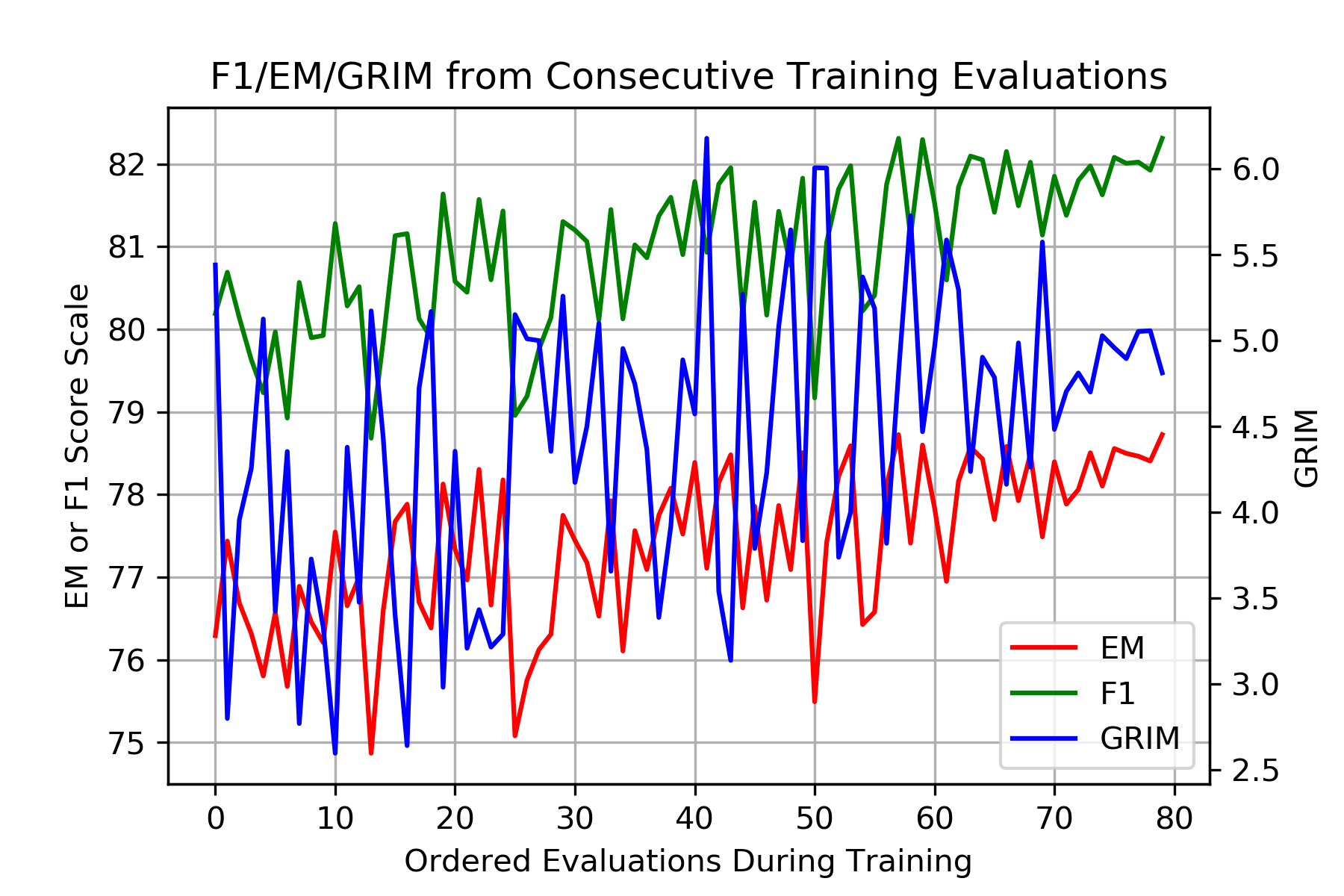}
\end{tabular}
\caption{Training progress assessed through regular validation runs for two experiments on a Roberta model(left) and a Bert model (right)}
\label{M:EM-GRIM}
\end{table*}

The high points of the oscillations of the F1 score occasionally establish a new ``best model" that defines a plateau until some other oscillation pushes the F1 value even higher, and a new best model plateau emerges. A  low GRIM value often coincides with a new best model. Since the magnitude of the oscillations tends to diminish, the GRIM values corresponding to each new best model get higher and higher, following another pattern of increasing plateaus. The evolution of these plateaus correlates to our informal notion of training maturity. Comparisons within similar levels of training maturity tend to show lower GRIM for higher F1s. However, when comparing models of different maturity, the plateau differences overwhelm the more subtle GRIM-performance relationship.

%% file: 45A_DirectPerformance.tex
\subsection{Ensembles based on F1, EM or GRIM}

Although the GRIM is based on secondary predictions, we were still interested to see if there is a more direct way to use it to improve performance. Here, we compare ensembles of models chosen based on GRIM, F1, and EM. The results show that the GRIM is a poor criterion. 

\begin{table}[ht]
\centering
\begin{tabular}{lrrl}
\toprule
 Ensemble Choice  &    EM  &  F1   & Models Selected (names abbreviated) \\
\midrule
     Best EM (baseline)  &   80.91  &	83.88 & bert-lg-uc\\
     Best 3 by EM  &  83.17 & 87.06 &  bert-lg-uc, bert-lg-cas, roberta/002 \\
    Best 3 by F1    &    83.08  & 87.30 &   bert-lg-uc, bert-lg-cas, roberta/001\\
 Best 3 by GRIM    &   81.11 & 85.49 & roberta, bert-lg-uc, bert/010 \\
     Best 5 by EM  &  83.34 & 87.21 &  bert-lg-uc, bert-lg-cas, roberta/002, longformer/011,roberta/001 \\
    Best 5 by F1    &    83.99  & 87.94 &   bert-lg-uc, bert-lg-cas, roberta/001, longformer/011, electra/006\\
 Best 5 by GRIM    &   81.76 & 85.73 & roberta, bert-lg-uc, bert/010, bert-lg-cas, 
 bert/009 \\
\bottomrule
\end{tabular}
\caption{Ensembles whose members were selected based on highest performance according to best EM, F1, or GRIM. The best single model is also shown for comparison.
}
\label{ensembleresults}
\end{table}

We used  the original 16 experiment results and tried various majority-vote ensembles with members determined by the best-3 and best-5 models with respect to F1, EM, or GRIM. The 5-way ensembles performed a little better than the respective 3-way. The results are in table \ref{ensembleresults}. We include the singleton best EM model for comparison. The Ensemble of the top-5 F1 scores performed the best, followed by the one selected based on the top-5 EM scores.

%% file: 70A_Discussion.tex
\section{Discussion and Future Work}


The Golden Rank (GR) at the individual example level, and the Golden Rank Interpolated Median (GRIM) at the aggregate level, quantify the quality of learning exhibited by a model in a way that parallels the very mechanism that produces the prediction. We have shown that these metrics can classify learning behavior through secondary predictions. We can group examples and isolate characteristics such as learning polarization, degree of confidence, and persistent failures from the outcomes of a few experiments. The resulting clusters can, in turn, expose patterns of failure and inconsistencies to aid the detection of errors and missing or unexpected golden answers.    

There were two considerations in the definition of these metrics that may cause criticism and  that we plan to further research:
\begin{enumerate}
    \item Our methodology opted to exclude primary predictions from the calculation of the GRIM. So we ended up with a secondary prediction classifier that is relatively sensitive and independent of EM and F1. But attributing a direction of ``goodness" became more complicated. Is it better for a model to exhibit low GRIM, indicating that secondary examples lean towards rank 0? Or is it better when the GRIM is higher, where all low-hanging fruit is pulled into rank 0, and the more challenging secondary examples remain in higher ranks? We introduced the informal notion of training maturity to qualify sufficiently the direction of goodness, which requires a better-informed definition. 
    One direction is to include the primary examples in the GRIM calculation, leading to a much narrower GRIM range but with a cleaner notion of the direction of ``goodness" and possibly more overlap with the EM and F1. Another is to keep the GRIM over secondary predictions and continue refining the quantification and impact of training maturity. 
    \item Another decision we made, at least at this phase, was to calculate the GR scores from the best-$K$ output of the evaluations. The overhead cost of computing the GR as part of the evaluation was a significant consideration initially, but it limited our ability  to calculate the actual value. Skipping the catch-all threshold K gimmick will allow us to use the unbiased mean, instead of the median, as the aggregate metric. In that scenario, the outliers with extreme values of GR (possibly in the hundreds or thousands) could make a more compelling case for discounting or normalization similar to other rank metrics used in web searches. It can also facilitate more direct ways of engineering the loss function to address class imbalance or easy negative example saturation. 
\end{enumerate}

We believe that the following areas need more investigation. 

\begin{description}
\item[Failed answer analysis] can bring further insight into how these metrics correlate with actual example failure patterns,  their association with linguistic patterns, or question types sensitive to failing on particular model architectures.

\item[To what extent can the GRIM aid in model selection?]  More data collection will likely provide insight into how different model architectures transition secondary predictions into primary "correct" predictions. Such understanding may lead to empirical indicators for controlling early stopping, as well as for architecture and hyperparameter selection.
\item[Explore double dip analysis.] We want to replicate some of the scenarios associated with the double-dipping phenomena observed during model training per \cite{nakkiran2019deep} and to see if any additional insight can be derived using the GR and GRIM, particularly during training evaluations. The notion of training maturity (which could also be cyclic instead of monotonic or random) seems quite relevant to this topic.

\item[Extend the GR and GRIM to other tasks and models.] Developing a metric class that can be included in the validation code will enable more data collection for different tasks than QA or outside the NLP realm. As a start, we intend to create a metric class for the Hugging Face QA examples. 
  
\item[Use GR score for adversarial  training:] We biased the dataset sampling for subsequent epochs to low (or high) GR examples. We tracked
the movement of examples that switched rank. We observed that equilibrium was reached, and as many examples moved towards lower ranks, as did in the opposite direction. Our findings were preliminary and are not included in this paper. Still, additional study is needed to understand this behavior better and potentially exploit it for adversarial training strategies.

\end{description}

%% file: 35A_Models.tex
\section{Models}\label{sec:models}

\subsection{BERT Models}

We use two implementations of BERT \cite{vaswani2017attention} and \cite{devlin2019bert} from Huggingface:  \href{https://huggingface.co/bert-base-uncased}{bert-base-uncased} with 12 layers and \href{https://huggingface.co/bert-large-cased-whole-word-masking}{bert-large-cased-whole-word-masking} with 24 layers. These models are pre-trained but not fine-tuned.  We fine-tuned three BERT-base experiments for eight and one for 16 epochs using various learning rates. We fine-tuned a BERT-large model for two epochs.

We also fine-tuned a ``BERT-large-cased-whole-word-masking" model for six epochs and  collected metrics from validations performed every 2000 optimizer steps to study GRIM behavior during training.

We finally evaluated a different large BERT model, already fine-tuned on SQuAD v2, \href{https://huggingface.co/deepset/bert-large-uncased-whole-word-masking-squad2}{deepset/bert-large-uncased-whole-word-masking-squad2} with the published weights.

\subsection{RoBERTa Models}

The acronym stands for \emph{A Robustly Optimized BERT Pre-training Approach} \cite{liu2019roberta}, also available in base and large configurations. As its name indicates, the focus is on a more robust method of pre-training BERT, performed on a broader corpus of the English language totaling  160GB of uncompressed text. The developers paid particular attention to optimizing the learning rate, momentum, and scheduling hyperparameters, along with tweaks to the training strategy. These tweaks entail dynamic masking, eliminating next sentence prediction,
training with large mini-batches and adopting a universal tokenization scheme based on byte-pair encoding.   

We used the \href{https://huggingface.co/navteca/roberta-base-squad2}{navteca/roberta-base-squad2} model, already fine-tuned on SQuAD v2.
We also used the pre-trained (but not fine-tuned)  \href{https://huggingface.co/roberta-base}{roberta-base} model \cite{10.1145/233269.233324}. We fine-tuned it on SQUAD-2 in a special experiment for over 16 epochs to collect metrics at each intermediate validation to understand how GRIM, EM and F1 evolve during training.

\subsection{ELECTRA Models}

These models also come in small and large configurations and are discussed in \cite{clark2020electra}. BERT randomly masks about 15\% of the incoming words and trains a decoder to recover them. In contrast, ELECTRA uses a simple generative model as a generator and trains a discriminator. The generator replaces the randomly selected words with plausible alternatives, and the discriminator detects the substituted words. Since the replacements have meaningful embeddings, the replaced tokens productively participate in the attention process. In contrast to the masked method, this approach results in faster and more economical training. 

We ran an evaluation with the published weights and two experiments further fine-tuning the \href{https://huggingface.co/navteca/electra-base-squad2}{navteca/electra-base-squad2} model from the Huggingface Hub. Details are shown in table \ref{tab:BERT-models}

\subsection{DistilBERT Models}

DistilBERT uses knowledge distillation, a model compression technique proposed by \cite{Bucila2006ModelC} that uses a teacher-student model. In this case, BERT is the teacher, and DistilBERT is the student. The student uses the logit output vector generated by the teacher as input to its training and learns to replicate the teacher's behavior with a reduced-size architecture. 
The resulting model is  a smaller and faster version of BERT, trained on the same corpus, that can be  used as a base for further task-specific fine-tuning to run on smaller platforms. It is described in \cite{sanh2020distilbert} where the authors claim that this model, which is 40\% smaller than BERT, retains 97\% of the original BERT understanding capabilities, and its training is  60\% shorter.  We used a fine-tuned implementation on SQuAD v2,  \href{https://huggingface.co/twmkn9/distilbert-base-uncased-squad2}{twmkn9/distilbert-base-uncased-squad2} from the Huggingface Hub, but were unable to come close to the performance of the other models as we show in our analysis.

We ran a validation with the published weights and two experiments extending the fine-tuning of this already fine-tuned Distilbert base model. 

\subsection{Longformer}

As the name implies, Longformer \cite{beltagy2020longformer} accommodates longer input sequences than BERT. To achieve this, the model restricts attention to a fixed window extending on either side of the attending token, essentially a  diagonal band matrix called windowed attention. As a result, attention cost is reduced from  quadratic to linear since the window is of fixed size. In addition, some critical tokens can  selectively participate in global attention, preserving their ability to exploit long-distance attention. For question-answering,  it is typical to set the question tokens to be global.

We ran a validation  on \href{https://huggingface.co/mrm8488/longformer-base-4096-finetuned-squadv2}{mrm8488/longformer-base-4096-finetuned-squadv2}  with the published weights. We then  extended its fine-tuning by a few epochs and reran a validation.

%% file: 37A_Metrics.tex
\section{ QA Post-processing and Metrics} \label{L2P}
We provide some context, more formally define our terminology, i.e., primary vs. secondary predictions and ranks, and review the post-processing steps of the QA model's outputs.

\subsection{From Logits to Predictions}

The rationalization of the Golden Rank draws from the way spans are prioritized and selected in classification \cite{NLPSbook}.
We assume that our evaluation set contains $M$ question-answer examples with varying context token sequence length $N^{(i)}$. For each example $i \in [1,M]$ the output layer of the model returns a start and an end prediction vector of values $\hat{y}^{S(i)}_j$  and $\hat{y}^{E(i)}_j$ where $j \in [1,N^{(i)}]$. We treat these two as logit vectors representing estimates of the  \emph{log of the odds} that the token in position $j$ marks the start or end of the answer segment, respectively.  These logit vectors are converted to prediction probability distributions $\hat{p}_j^{S(i)}$ and $\hat{p}_j^{E(i)}$ by applying the softmax function. 

 In a perfect world, we should be able to use the highest probability start and end indexes to identify the best answer prediction. 
    $$S^{(i)}=\underset{j\in [1,N]}{\arg\max}\,(\hat{p}_j^{S(i)})$$
    $$E^{(i)}=\underset{k\in[1,N]}{\arg\max}\,(\hat{p}_k^{E(i)})$$
This best answer prediction is the span of the context that starts at position $S^{(i)}$ and ends at $E^{(i)}$.   
Before accepting this result, however, we must ensure that these answer spans are valid. For example, the end position cannot precede the start position, the number of tokens in the span cannot exceed a threshold length, or either position cannot be outside the context bounds. If the highest probability span is invalid, we need a strategy for picking the next candidate. Should we try the next best end, maybe the next start, or skip to the next pair? Or should we consider the question unanswerable?  
 
Instead of dealing with two independent probability distributions for the starts and ends, it would be better to have a composition function that generates an ordered list in descending probability order for each possible span. 

Visualize a probability matrix of all possible combinations of start positions (as row indexes) and end positions (as column indexes): $\bf{\hat{p}}^{(i)} \in \mathcal{R}^{N^{(i)} \times N^{(i)}}$. Each matrix element represents a possible span and contains the probability that the span matches a correct answer. Elements denoting invalid spans must have their probability set to zero. For example, all elements below the diagonal that denote ends before starts should be zero, as should those outside the band where the answer span length exceeds the threshold of allowable answer length. Depending on how these validity rules are applied, rows may  not add up to one. Since these values are only used to  ``rank" predictions, the fact that they are not strict probabilities does not alter the ordering. Consequently, we don't need to apply another softmax. 

Returning to the composition function, we want it to return a single probability per start-end pair derived from the start and end probabilities. To compute the combined probability, we can take the product of the start and end probabilities while enforcing span validity rules:
\begin{equation}
\hat{p}^{(i)}_{j,k}  =  \left\{
    \begin{array}{ll}
    \hat{p}_j^{S(i)} \hat{p}_k^{E(i)} & \text{for }0 \leq k-j \leq L\\
    0   &   \text{otherwise}
    \end{array}
        \right.
\label{phat-jk}
\end{equation} 

$L$ is the maximum answer length in tokens.  

To avoid multiple applications of the softmax, separately for start and end, we apply it once over the sum of the logits, which  turns out to be equivalent since:

\begin{align*}
     \hat{p}_j^{S(i)} \hat{p}_k^{E(i)} & = \frac{\exp{\hat{y}^{S(i)}_j} \exp{\hat{y}^{E(i)}_k}}{\sum_{q=1}^n \exp{\hat{y}^{S(i)}_q} \sum_{r=1}^n \exp{\hat{y}^{E(i)}_r}} 
       & \\
     & =  \frac{\exp\left({\hat{y}^{S(i)}_j + \hat{y}^{E(i)}_k}\right)}
     {\sum_{q,r=1}^n \exp\left({\hat{y}^{S(i)}_q + \hat{y}^{E(i)}_r}\right) }
\end{align*}

It is also possible to apply different weights to the start and end. The idea here is to emphasize the probability of the starting token more than the ending token, as discussed in \cite{SchwagerQABERT}. The last equation looks like this: 
   \[ \hat{p}^{(i)}_{j,k} =
   \frac{\exp\left({w^S \hat{y}^{S(i)}_j + w^E \hat{y}^{E(i)}_k}\right)}
     {\sum_{q,r=1}^n \exp\left({w^S\hat{y}^{S(i)}_q + w^E\hat{y}^{E(i)}_r}\right) } \]

%% file: A20_Examples.tex
\section{Error Analysis - 4 Experiments at (1,0)}
\label{sec:A21}

The four  examples in this appendix have GR = 1 for all experiments (and a standard deviation of 0.) We consider the coincidence worth investigating. The top two ranks predicted by one of our BERT-large models are shown in table \ref{4QuestionsTab}.

In summary, three of the four examples produce a ``No-Answer"  rank-0 prediction, while all four have their second choice match a correct answer. 

 The last question seems to be phrased wrong. It is stated as ``residential construction can generate what \emph{is} not carefully planned?" while it should be  ``residential construction can generate what \emph{if} not carefully planned?"

The four questions follow:

\subsection{Example ID: 57263c78ec44d21400f3dc7c}

This  example has a spelling error in the question (between) which may have caused the model to attribute a higher probability to the no-answer.

\begin{description}
    
\item[title:] Packet switching

\item[context:] ARPANET and SITA HLN became operational in 1969. Before the introduction of X.25 in 1973, about twenty different network technologies had been developed. Two fundamental differences involved the division of functions and tasks between the hosts at the edge of the network and the network core. In the datagram system, the hosts have the responsibility to ensure orderly delivery of packets. The User Datagram Protocol (UDP) is an example of a datagram protocol. In the virtual call system, the network guarantees sequenced delivery of data to the host. This results in a simpler host interface with less functionality than in the datagram model. The X.25 protocol suite uses this network type.

 \item[question:] 2 differences betwen X.25 and ARPNET CITA technologies.

 \item[answers:] \hspace{10pt}
    \begin{itemize}
    \item  Two fundamental differences involved the division of functions and tasks between the hosts at the edge of the network and the network core.
    \item the division of functions and tasks between the hosts at the edge of the network and the network core.
   \item division of functions and tasks between the hosts at the edge of the network and the network core',
  \end{itemize}
 
 \item[answer start:] 154, 191, 195
 \end{description}
 
\subsection{Example ID: 57267d52708984140094c7da}

The rank-0 answer for this example is ``microscopic analysis of oriented thin sections of geologic samples", which appears to be a correct answer as it appears, but it is not included in the list of golden answers. There is an identical correct answer preceded by the verb ``uses", but given the pattern of the other two correct answers excluding the verb should also be accepted as correct. In this case the F1 test would have fairly evaluated a close answer. 

\begin{description}
    
\item[title:] Geology
\item[context:] Structural geologists use microscopic analysis of oriented thin sections of geologic samples to observe the fabric within the rocks which gives information about strain within the crystalline structure of the rocks. They also plot and combine measurements of geological structures in order to better understand the orientations of faults and folds in order to reconstruct the history of rock deformation in the area. In addition, they perform analog and numerical experiments of rock deformation in large and small settings.

 \item[question:] How do structural geologists observe the fabric within the rocks?

 \item[answers:] \hspace{10pt}
    \begin{itemize}
    \item microscopic analysis of oriented thin sections
    \item microscopic analysis
    \item use microscopic analysis of oriented thin sections of geologic samples
    \end{itemize}
 
 \item[answer start:] 26, 26, 22
 \end{description}

\subsection{Example ID: 5728dc2d3acd2414000e0080}

This appears to be a learning/generalization problem and does not show any apparent traces of why all models fail to recognize the correct answer, although they come so close as to rank it the second-best answer.

\begin{description}
    
\item[title:] Civil disobedience
\item[context:] Some theories of civil disobedience hold that civil disobedience is only justified against governmental entities. Brownlee argues that disobedience in opposition to the decisions of non-governmental agencies such as trade unions, banks, and private universities can be justified if it reflects "a larger challenge to the legal system that permits those decisions to be taken". The same principle, she argues, applies to breaches of law in protest against international organizations and foreign governments.

 \item[question:] Who claims that public companies can also be part of civil disobedience?

 \item[answers:] \hspace{10pt}
    \begin{itemize}
    \item Brownlee
    \item Brownlee
    \item Brownlee
    \item Brownlee
    \item Brownlee
    \end{itemize}
 
 \item[answer start:] 114, 114, 114, 114, 114
 \end{description}

\subsection{Example ID: 572742bd5951b619008f8787}

This question is phrased incorrectly; actually, it is incorrect English. The proper way is to replace the word \emph{is} by \emph{if} be: ``Residential construction can generate what if not carefully planned?"

\begin{description}
    
\item[title:] Construction
\item[context:] Residential construction practices, technologies, and resources must conform to local building authority regulations and codes of practice. Materials readily available in the area generally dictate the construction materials used (e.g. brick versus stone, versus timber). Cost of construction on a per square meter (or per square foot) basis for houses can vary dramatically based on site conditions, local regulations, economies of scale (custom designed homes are often more expensive to build) and the availability of skilled tradespeople. As residential construction (as well as all other types of construction) can generate a lot of waste, careful planning again is needed here.

 \item[question:] Residential construction can generate what is not carefully planned?

 \item[answers:] \hspace{10pt}
    \begin{itemize}
    \item a lot of waste
    \item waste
    \item waste
    \end{itemize}
 
 \item[answer start:] 629, 638, 638
 \end{description}

%% file: A21_ExamplePredictions4.tex
\begin{table}
\caption{Top two predictions from model bert-large-cased-whole-word-masking/003 on each of the four questions. Only the last four characters of the id are shown.}
\label{4QuestionsTab}
\begin{tabular}{p{6cm}rlrp{10cm}l}
\toprule
    prediction &  probability &   id &  rank &  goldAns &  correct \\
\midrule
{} & 9.999964e-01 & dc7c & 0 & [Two fundamental differences involved the division of functions and tasks between the hosts at the edge of the network and the network core, 

the division of functions and tasks between the hosts at the edge of the network and the network core., 

division of functions and tasks between the hosts at the edge of the network and the network core] & False \\
Two fundamental differences involved the division of functions and tasks between the hosts at the edge of the network and the network core & 8.562545e-07 & dc7c & 1 & [Two fundamental differences involved the division of functions and tasks between the hosts at the edge of the network and the network core, 

the division of functions and tasks between the hosts at the edge of the network and the network core., 

division of functions and tasks between the hosts at the edge of the network and the network core] & True \\
microscopic analysis of oriented thin sections of geologic samples & 8.835049e-01 & c7da & 0 & [microscopic analysis, 

microscopic analysis of oriented thin sections, 

use microscopic analysis of oriented thin sections of geologic samples] & False \\
use microscopic analysis of oriented thin sections of geologic samples & 9.161043e-02 & c7da & 1 & [microscopic analysis, 

microscopic analysis of oriented thin sections, 

use microscopic analysis of oriented thin sections of geologic samples] & True \\
{} & 9.999518e-01 & 0080 & 0 & [Brownlee] & False \\
Brownlee & 4.670037e-05 & 0080 & 1 & [Brownlee] & True \\
{} & 9.999995e-01 & 8787 & 0 & [waste, a lot of waste] & False \\
waste & 3.547627e-07 & 8787 & 1 & [waste, a lot of waste] & True \\
\bottomrule
\end{tabular}
\end{table}

%% file: A30_PolarizedExps.tex
\section{Error Analysis - Highly Polarized Examples}
\label{sec:A30}

Highly polarized examples have some experiments rank them very low and some very high.
The highest standard deviation appears with unanswerable examples where the model thought there was an answer. Longer answers  often come in  clusters overlapping sub-spans forming plausible answers in a sequence that pushes the next answer cluster, or a ``no answer", towards higher ranks, frequently falling into the 10+ bucket. So the GR for models that predicted correctly is zero, while for those that didn't may be much higher, making the example look polarized and volatile.  

The examples below are such highly volatile cases. The following table shows results from the second example, including the rank 0, rank 1, and the correct answer if its rank is higher. As a result, we get a flavor of how different models answer such questions and how they end up in high ranks. The two large BERT models, RoBERTa, Longformer, and Electra, tend to answer these questions correctly (rank-0), while others, such as BERT-base and DistilBERT, do not.

Tuning the Hugging Face argument \emph{no answer probability threshold} as shown in \cite{HF_squad2_eval} may improve the success rate of these examples.  

\subsection{Example ID: 5ad56bcd5b96ef001a10ae62}

\begin{description}
\item[title:] Computational complexity theory
\item[context:] Many known complexity classes are suspected to be unequal, but this has not been proved. For instance $P \subseteq NP \subseteq PP \subseteq PSPACE$, but it is possible that $P = PSPACE$. If P is not equal to NP, then P is not equal to PSPACE either. Since there are many known complexity classes between P and PSPACE, such as RP, BPP, PP, BQP, MA, PH, etc., it is possible that all these complexity classes collapse to one class. Proving that any of these classes are unequal would be a major breakthrough in complexity theory.

 \item[question:] What is the proven assumption generally ascribed to the value of complexity classes?

 \item[answers:] No Answer
 
 \item[answer start:] 0
 \end{description}
 
 \subsection{Example ID: 5ad251d6d7d075001a428ceb}

\begin{description}
\item[title:] Oxygen
\item[context:] The unusually high concentration of oxygen gas on Earth is the result of the oxygen cycle. This biogeochemical cycle describes the movement of oxygen within and between its three main reservoirs on Earth: the atmosphere, the biosphere, and the lithosphere. The main driving factor of the oxygen cycle is photosynthesis, which is responsible for modern Earth's atmosphere. Photosynthesis releases oxygen into the atmosphere, while respiration and decay remove it from the atmosphere. In the present equilibrium, production and consumption occur at the same rate of roughly 1/2000th of the entire atmospheric oxygen per year.

 \item[question:] What oxygen reservoir is the driving factor of the oxygen cycle?

 \item[answers:] No Answer
 
 \item[answer start:] 0
 \end{description}

%% file: A32_PolarizedExps.tex
\begin{table}
\begin{tabular}{p{5cm}rlrll}
\toprule
                                                         text &  probability &             experiment &  rank & goldAns &  correct \\
\midrule
photosynthesis & 9.999955e-01 & bert-uc/001 & 0 & [] & False \\
photosynthesis, & 2.932970e-06 & bert-uc/001 & 1 & [] & False \\
{} & 5.623741e-14 & bert-uc/001 & 10 & [] & True \\
photosynthesis & 9.813934e-01 & bert-uc/009 & 0 & [] & False \\
The main driving factor of the oxygen cycle is photosynthesis & 5.801043e-03 & bert-uc/009 & 1 & [] & False \\
{} & 2.939046e-07 & bert-uc/009 & 86 & [] & True \\
photosynthesis & 9.619037e-01 & bert-uc/010 & 0 & [] & False \\
The main driving factor of the oxygen cycle is photosynthesis & 2.637961e-02 & bert-uc/010 & 1 & [] & False \\
{} & 8.520884e-06 & bert-uc/010 & 10 & [] & True \\
{} & 7.064449e-01 & bert-lg-cas-wwm/003 & 0 & [] & True \\
the atmosphere, the biosphere, and the lithosphere & 1.259139e-01 & bert-lg-cas-wwm/003 & 1 & [] & False \\
{} & 9.994391e-01 & bert-lg-uc-wwm-sq2/001 & 0 & [] & True \\
photosynthesis & 1.410887e-04 & bert-lg-uc-wwm-sq2/001 & 1 & [] & False \\
photosynthesis & 9.997396e-01 & longformer-sq2/009 & 0 & [] & False \\
photosynthesis, & 1.777181e-04 & longformer-sq2/009 & 1 & [] & False \\
{} & 1.237061e-08 & longformer-sq2/009 & 10 & [] & True \\
{} & 5.466675e-01 & longformer-sq2/011 & 0 & [] & True \\
photosynthesis & 4.083973e-01 & longformer-sq2/011 & 1 & [] & False \\
{} & 9.999925e-01 & electra-sq2/005 & 0 & [] & True \\
photosynthesis & 1.891310e-06 & electra-sq2/005 & 1 & [] & False \\
{} & 9.999998e-01 & electra-sq2/007 & 0 & [] & True \\
photosynthesis & 3.729614e-08 & electra-sq2/007 & 1 & [] & False \\
photosynthesis & 9.300030e-01 & roberta-sq2/001 & 0 & [] & False \\
photosynthesis, & 6.051588e-02 & roberta-sq2/001 & 1 & [] & False \\
{} & 2.919881e-06 & roberta-sq2/001 & 10 & [] & True \\
{} & 5.483240e-01 & roberta-sq2/002 & 0 & [] & True \\
the atmosphere, the biosphere, and the lithosphere & 8.345626e-02 & roberta-sq2/002 & 1 & [] & False \\
photosynthesis & 9.972602e-01 & distilbert-uc-sq2/003 & 0 & [] & False \\
photosynthesis, & 1.422049e-03 & distilbert-uc-sq2/003 & 1 & [] & False \\
{} & 2.097861e-12 & distilbert-uc-sq2/003 & 10 & [] & True \\
\bottomrule
\end{tabular}
\caption{Selective GR ranks from various experiments for example ad251d6d7d075001a428ceb}
\end{table}